\documentclass[preprint,12pt]{elsarticle}

\graphicspath{ {./figures/} }
\usepackage{hyperref}
\usepackage{float}
\usepackage{verbatim} 
\usepackage{apalike}

\usepackage{mathtools}
\usepackage{amsmath} 
\usepackage{algorithm}
\usepackage{algpseudocode}
\usepackage{multirow}

\usepackage{xcolor}
\usepackage{soul}

\definecolor{aqua}{rgb}{0.0, 1.0, 1.0}

\usepackage{amssymb}

\biboptions{authoryear}

\begin{document}
\begin{frontmatter}

\begin{titlepage}
\begin{center}
\vspace*{1cm}

\textbf{ \large Dynamic Task and Weight Prioritization Curriculum Learning for Multimodal Imagery}

\vspace{1.5cm}

Hüseyin Fuat Alsan $^a$ (huseyinfuat.alsan@stu.khas.edu.tr), Taner Arsan$^a$ (arsan@khas.edu.tr) \\

\hspace{10pt}

\begin{flushleft}
\small  
$^a$ Computer Engineering Department, Kadir Has University, Istanbul, Turkey \\

\vspace{1cm}
\textbf{Corresponding Author:} \\
Hüseyin Fuat Alsan \\
Computer Engineering Department, Kadir Has University, Istanbul, Turkey \\
Tel: (+90) 0536 825 29 11 \\
Email: huseyinfuat.alsan@stu.khas.edu.tr

\end{flushleft}        
\end{center}
\end{titlepage}

\begin{abstract}
This paper explores post-disaster analytics using multimodal deep learning models trained with curriculum learning method. Studying post-disaster analytics is important as it plays a crucial role in mitigating the impact of disasters by providing timely and accurate insights into the extent of damage and the allocation of resources. We propose a curriculum learning strategy to enhance the performance of multimodal deep learning models. Curriculum learning emulates the progressive learning sequence in human education by training deep learning models on increasingly complex data. Our primary objective is to develop a curriculum-trained multimodal deep learning model, with a particular focus on visual question answering (VQA) capable of jointly processing image and text data, in conjunction with semantic segmentation for disaster analytics using the FloodNet dataset. To achieve this, U-Net model is used for semantic segmentation and image encoding. A custom built text classifier is used for visual question answering. Existing curriculum learning methods rely on manually defined difficulty functions. We introduce a novel curriculum learning approach termed Dynamic Task and Weight Prioritization (DATWEP), which leverages a gradient-based method to automatically decide task difficulty during curriculum learning training, thereby eliminating the need for explicit difficulty computation. The integration of DATWEP into our multimodal model shows improvement on VQA performance. Source code is available at https://github.com/fualsan/DATWEP.
\end{abstract}


\end{frontmatter}

\section{Introduction}
\label{introduction}

 Multimodal deep learning is about using (and relating between) different data types and can include images, text, audio, time series data and even tabular data. It can be seen as a generalization method for deep learning at the data level. Relations between different types (modalities) can be automatically constructed via pattern recognition commonly employed by deep learning models. However, training with multimodal data can be challenging and often requires efficient training algorithms. Curriculum learning is considered in this work to increase multimodal performance. Instead of random shuffling the data samples, curriculum learning schedules training in a meaningful order, from the data samples that are easy for deep learning models learn to the data samples that are hard to learn. A difficulty measurement function is required to distinguish between easy and hard data samples, and a pacing function is for scheduling the speed of introducing hard samples. Using multimodal data and curriculum learning, deep learning models mimic how humans learn, increasing training efficiency. Multimodal curriculum learning can be applied to various deep learning scenarios. In this work, we focus on flooding disaster analytics with aerial imagery where image and text data are used.

Flooding is one of the most important disasters; globally, accounts for 44\% of all disasters and significantly affected around 1.6 billion people between 2000 and 2019 \citep{un_disaster_report}. Furthermore, sea levels are expected to rise in upcoming years between 2020 and 2050, increasing the probability of flooding even more \citep{nos_sea_level_report}. So flooding should be taken very seriously since it can significantly cause ecological, environmental, and social security problems. It is predicted that the sea level rise will threaten around 140 to 170 million people until the year 2050, and approximately 310 to 420 million people will be threatened until the year 2100 \citep{kulp_disaster_coastal_flooding}. Rapid rescue efforts and quick flood detection can help with recovery. Most importantly, collected aerial imagery data around the disaster zones can be quickly analyzed in detail. Today, aerial imagery can be collected with two methods: satellites or unmanned aerial vehicles (UAVs) like drones with less human effort. UAVs may readily access challenging sites and take high-resolution pictures that display the state of the affected surroundings.

\bigskip

This paper uses the FloodNet\footnote{https://github.com/BinaLab/FloodNet-Challenge-EARTHVISION2021} \citep{rahnemoonfar_floodnet_original} dataset, which offers high-resolution images shot from low altitudes with semantic segmentation and visual question-answering annotations. Having two tasks at hand allows for an in-depth analysis of disasters. The semantic segmentation track helps identify the different objects, their shape, and their position in the image. On the other hand, visual question answering (VQA) can answer arbitrary questions about object and scene properties, such as condition recognition, attribute classification, counting, etc., can be very helpful in damage assessment predictions. Using high-accuracy visual question answering systems to detect and comprehend floods. Also, images are high resolution, and scenes are clean because FloodNet's collected data can assist deep learning models in determining post-disaster damage assessment decisions with greater accuracy.

By leveraging deep learning algorithms, the post-disaster analysis process becomes more efficient, significantly reducing the reliance on human labor. Semantic segmentation accurately identifies damaged areas and collapsed structures, while real-time monitoring of high-resolution imagery ensures up-to-date situational awareness. Semantic segmentation techniques can accurately delineate objects and regions of interest within the images, allowing for a precise assessment of the extent of the disaster. VQA helps with gathering more information about the images by asking questions. Combining segmentation and VQA, we can develop information retrieval methods that can aid in resource allocation and prioritization of response efforts.

We aim to train multimodal deep learning models for post-disaster analytics efficiently. With the FloodNet dataset, we have two tasks: semantic segmentation, which contains image data, and visual question answering, which includes text data. So we formulated the learning schema as multitask, multimodal deep learning. A gradient-based curriculum learning approach is proposed for training multi-task and multimodal deep learning models. Our contributions are:

%

\begin{itemize}
\item Dynamic task prioritization between segmentation and visual question answering method (called DATAP) where the model chooses the task to focus on by minimizing the gradients w.r.t the current total loss.

\item Dynamic weight prioritization for answer classes in visual question answering task method (called DAWEP) where the model itself chooses which answer classes to focus on by minimizing the gradients w.r.t the current VQA loss.

\item Combining those above, we present a novel curriculum learning-based approach for dynamically adjusting task priority and answering class weights during training (called DATWEP). DATWEP is gradient based curriculum method which does not  require difficulty function to be explicitly defined. 

\item We implement a multimodal model for handling both semantic segmentation and visual question-answering tracks of FloodNet at the same time instead of implementing separate models for each track. Using DATWEP, we obtain a comparable performance to the FloodNet baselines using only $\sim$9.5M parameter multimodal deep neural network.

\end{itemize}

This paper is organized as follows: Section \ref{sec_related_works} discusses similar works in the literature. Section \ref{sec_methodology} explains preprocessing of the FloodNet dataset, the architecture of our deep learning model, loss and evaluation metrics and our contributing curriculum learning method. Section \ref{sec_experiments} describes how experiments are conducted, discusses our results and compares our results with other baseline FloodNet models. Finally, Section \ref{sec_conclusions_futurework} summarizes our work.

\section{Related Works}\label{sec_related_works}
\subsection{Semantic Segmentation}

Image segmentation is one of the main topics of study in computer vision and is very useful for comprehending scenes. Segmentation partitions the given image into different regions and their boundaries. It is mainly used for locating different objects in an image. Nowadays, segmentation is accomplished with deep neural network models where the output of the model is a pixel map, called segmentation mask where object regions and boundaries are pixel coded. Segmentation is flexible since segmentation masks can take any arbitrary shape. In this work, we focus on semantic segmentation, a subfield of segmentation where the
mask of each class object is assigned to a label.

Convolutional neural networks (CNNs) are the most widely used architecture in segmentation (and computer vision in general). CNNs are introduced in \citep{fukushima_seg_first_conv, waibel_seg_first_conv2} and improved with backpropagation with \citep{lecun_seg_first_conv_backprop}. Fully convolutional networks (FCN) \citep{long_seg_fcn} are particularly valuable for segmentation because they output an image that can be used as segmentation mask. Segmentation models can be further classified into several sub-categories: Encoder-Decoder, Pyramid, and Regional CNN (R-CNN) based networks.

E-Net is an encoder-decoder \citep{paszke_enet} network that uses two processing stages. First input images are compressed into a latent representation (encoding) and upsampled from latent to original dimensions (decoding). U-Net \citep{ronneberger_seg_unet} is another encode-decoder network with skip connections between the encoding and decoding part of the network so some features are passed into upsampling without modification. Feature Pyramid Network (FPN) \citep{lin_seg_fpn}, uses bottom-up and top-down approaches similar to encoder-decoder networks to extract multi-scale image features. PSPNet \citep{zhao_pyramid_scene_parsing} is a pyramid pooling based network that uses convolutions with different pool sizes to extract image features in different scales. Combining the advantages of encoder-decoder and pyramid pooling, DeepLabV3+ \citep{chen_encoder_decoder_sep_segmentation} uses dilated separable convolutions to improve segmentation. Mask R-CNN \citep{he_seg_mask_rcnn} is a modified version of the Faster R-CNN \citep{ren_seg_faster_rcnn} network for segmentation. R-CNN based networks use a region proposal network (RPN) that predicts region of interest (RoI) proposal areas in the given image. \citep{wang_oriental_attention} introduces the orientation attention network (OANet) as a solution to the challenge of accurate semantic segmentation in high-resolution remote sensing images characterized by complex textures and edges. The OANet employs an asymmetrical convolution (AC) to capture directional anisotropy and an orientation attention module (OAM) to enhance geometric features by selecting beneficial features along coordinate axes. Extracting scale information is studied by \citep{peng_mining_scale_information} by proposing a novel network called MSINet, which leverages digital surface models (DSMs) to efficiently capture scale information. MSINet accomplishes this through an interpolation pyramid algorithm that encodes scale information from DSMs, providing scale prior information to the segmentation network, and by incorporating a spatial information enhancement module and a mutual-guidance module to address noise in the DSM boundaries. \citep{yi_cascade_composite_transformer} introduces a cascade composite transformer-based semantic segmentation network (CCTseg) to tackle large-scale object size variations due to altitude and angle changes from UAV captured images. This network incorporates a cascade composite encoder, spatial-enhanced transformer blocks, and a symmetric rhombus decoder to effectively capture context and local information.

In our implementation, we use an encoder-decoder type network U-Net, trained with our curriculum learning method. The U-Net architecture is chosen for semantic segmentation due to its ability to capture both local and global features effectively, combining them for accurate object segmentation. Its symmetric U-shaped design with skip connections allows for precise localization, while dense predictions at the pixel level enable fine-grained segmentation. However, our main difference is that our model is trained in a multimodal fashion, along with the question classifier model.

\subsection{Visual Question Answering}

Visual question answering is a multimodal task that combines computer vision and natural language processing (NLP) and, thus, works with both image and text (question sentence) data. The main objective is to answer open-ended questions about a given image. In the most basic form, image and text data are encoded with separate deep neural network models to generate feature representations for both. In \cite{antol_vqa_original} image features are encoded with CNN, text features are encoded with Long short-term memory (LSTM) models and these representations are point-wise multiplied and passed on to another fully connected layer to predict answers. Another approach is to concatenate encoded features to get higher dimensional representations \citep{zhou_simple_baseline_vqa}. In our implementation, we also do concatenation, but since we're using U-Net, image features from different scales are used instead of single feature representations like in \cite{antol_vqa_original} and \cite{zhou_simple_baseline_vqa}.

Attention-based methods are given by \cite{yu_factorized_vqa} and \cite{yang_stacked_attention_vqa}, which improve the learning of image and text representations. In \cite{yu_factorized_vqa}, a co-attention method is used to jointly learn image and text features. A similar approach is proposed by \cite{yang_stacked_attention_vqa}, but attention is stacked across several layers to progressively answer questions. Transformer \citep{vaswani_vqa_transformer_original} is a commonly used attention based that was originally introduced for natural language processing tasks but is widely adopted in computer vision and multimodal tasks (like VQA). Vision and language transformer (ViLT) \citep{kim_vqa_vilt_original} is a model for VQA tasks without using traditional CNNs or recurrent layers. ViLT uses a transformer encoder layer where images are divided into smaller flattened patches, and words are tokenized to produce multimodal sequences of data. \citep{sarkar_sam_vqa} present the supervised attention-Based visual question answering (SAM-VQA) model, a framework for VQA. It utilizes a Resnet-152 architecture for image feature extraction and a two-layer LSTM for question feature extraction. The model employs MFB pooling to create a detailed multimodal representation and applies a softmax function to estimate attention weights for questions in relation to images.

Aerial imagery with VQA is studied by \citep{lobry_vqa_remote_sensing, zheng_vqa_mutual_attention_remote_sensing} with curriculum learning \citep{yuan_vqa_easytohard_cl_remote_sensing, yuan_vqa_selfpaced_cl_remote_sensing}. By \cite{yuan_vqa_easytohard_cl_remote_sensing}, question length is taken as prior knowledge of difficulty and dynamically adjusted to achieve more effective training of the VQA model. Similarly, in \citep{yuan_vqa_selfpaced_cl_remote_sensing} self-paced curriculum learning is used with hard and soft weighting methods for training the VQA model. Alternative to curriculum learning, in \cite{lobry_vqa_remote_sensing} point-wise multiplication of image and text features is used similarly to  \citep{antol_vqa_original}. In \cite{zheng_vqa_mutual_attention_remote_sensing} image and text features are fused in "mutual attention component" layer to obtain multimodal features and concatenated with the original image and text features and eventually fuse all features into a bilinear layer. Besides the Floodnet \citep{rahnemoonfar_floodnet_original} dataset, RSVQA dataset is also used for remote sensing applications of VQA \citep{lobry_rsvqa_original}. \citep{lowande_feasibility_vqa} combines an CNN based image classification model with a Bag of Words based language model to not only detect damage but also answer questions related to post-disaster scenarios.

Transformer model with remote sensing VQA is studied by \cite{chappuis_vqa_transformer_remote_sensing, alsaleh_vqa_transformer_open_ended_remote_sensing, bazi_vqa_transformer_bi_modal_remote_sensing}. In \citep{chappuis_vqa_transformer_remote_sensing}, the transformer model is used only for text (question sentence) encoder and a traditional CNN model encodes the image. However, in \cite{alsaleh_vqa_transformer_open_ended_remote_sensing} and  \cite{bazi_vqa_transformer_bi_modal_remote_sensing} both image and text are encoded by a transformer model with Vision Transformer (ViT) \citep{dosovitskiy_vit_original} as the image encoder. Image and text can be separately encoded and fused with a transformer decoder \citep{alsaleh_vqa_transformer_open_ended_remote_sensing} or encoded with contextual relationships between image and text representations \citep{bazi_vqa_transformer_bi_modal_remote_sensing}.

\subsection{Curriculum Learning}\label{subsec_lit_review_cl}
Curriculum learning (CL) is a learning procedure where the training data's complexity gradually increases during training instead of randomly choosing the training data. Since it closely resembles how people actually learn, this approach to train deep learning models is natural. This paper focuses on developing a dynamic curriculum at the task and class level (explained in Section \ref{sec_proposed_CL}). So our most significant contribution to the paper is in this area. Curriculum learning can be divided into several subfields: vanilla CL, self-paced learning (SPL), self-paced CL (SPCL), teacher-student CL, and implicit CL.

In the early works of vanilla curriculum learning, \citep{bengio_cl_entropy} suggested curriculum learning as a way to increase the entropy of training distributions and apply to geometric shapes and language modeling gradually increasing the number of complex geometric shapes and frequent words. In \cite{spitkovsky_cl_sentence_len}, sentence length is gradually increased as a measure of difficulty. Results in these initial methods were successful and gave the green light to CL. However, these methods require difficulty to be priorly known and hand-picked. Self-paced learning does not require difficulty to be known a priori and can be calculated during training. In this way, the difficulty is periodically calculated during training, changing the samples' order in the process \citep{kumar_cl_selfpacedlearning}. Self-paced Curriculum Learning combines both prior and measurement during training \citep{jiang_self_paced_curriculum_learning}. Self-paced learning is applied to object detection  \citep{soviany_spcl_crossdomain_objectdetection, zhang_spcl_weakly_object_detection}, semantic segmentation \citep{pan_sp_medical_segmentation_clustering, zhang_sp_xray_noise_labels, peng_sp_constrastive_meta_labels}, and visual question answering \citep{yuan_spcl_vqa_remote_sensing}.

Teacher-student CL is another method where a pretrained teacher model chooses the curriculum for a student model. This method was used by \cite{hacohen_cl_teacherstudent_initial} to lay foundational work on common CL concepts such as pacing and difficulty with different case studies. Later research, however, indicates that it is mostly used in reinforcement learning \citep{matiisen_cl_teacherstudent_cl}, \citep{bouri_cl_studentteacher_hospital, jiang_cl_mentornet}. The intuition behind these methods is that student models aren't that experienced in selecting the correct curriculum, so teacher models direct student models to choose between easy and hard curricula. It is analogous to the human education system because if a student knows little about the course, it makes sense to ask an experienced teacher to assist the student in evaluating the materials for the curriculum.

CL can also be applied implicitly. Using Gaussian kernels, curriculum learning is implicitly applied in \cite{sinha_cl_gaussian_smoothing}, reducing the alias and smoothing the feature maps of CNNs. Curriculum learning can also be applied to unsupervised domain adaption \citep{almeida_cl_unsupervised_domain_adaption} to minimize sample selection entropy. \citep{kesgin_cyclical_cl} implements cyclical curriculum learning (CCL) which cyclically changes the size of the training dataset rather than simply increasing or decreasing it. By alternating between curriculum and non-curriculum learning, CCL achieves more successful results when applied to various architectures and datasets for image and text classification tasks.

\subsection{Comparison of Proposed CL Method to Literature}
Our contribution is mostly related to self-paced learning. The original implementation \citep{kumar_cl_selfpacedlearning} of loss is given in Eq. \ref{equ_spl_original} where $N$ is the number of data samples (in current batch), $\textbf{w} = [w_1, w_2,\dots, w_M] \in \mathbb{R}^M $ is a vector of trainable parameters of the model with the size of $M$, $\textbf{v} = [v_1, v_2,\dots,v_N] $ $\in [0, 1]^N$ is vector of weights for each each loss value $l_i, \ i=1,2,\dots,N$ and $r(\textbf{v};\lambda)$ is the SP-regularizer given in Eq. \ref{equ_spl_original_g}. SP-regularizer introduces harder examples by adjusting the weights with the help of age parameter $\lambda$ as the training progresses. In SPL, easy examples are considered to have lower loss values whereas hard examples are considered to have higher loss values. The age parameter acts as threshold such that $v_i = 1$ for $l_i < \lambda$ and $v_i = 0$ for vice versa. Therefore, gradually increasing $\lambda$ parameter will introduce harder examples and the rate of increase will determine the pacing or learning. Loss values are calculated traditionally with any neural network using gradient descent and backpropagation. In summary, SPL applies a weighted loss term  to all training data samples and by changing the the age parameter $\lambda$, weights are adjusted during training. There are also different regularizers such as linear and logarithmic \citep{jiang_cl_spl_reg_lin_logarithmic}, logistic \citep{xu_cl_spl_reg_logistic} and polynomial \citep{gong_cl_spl_reg_polynomial}.

\begin{equation}\label{equ_spl_original}
	\footnotesize
	\min \limits_{\textbf{w}; \textbf{v} \in [0,1]^N} \mathcal{L}(\textbf{w}, \textbf{v}; \lambda) =  \sum_{i=1}^N v_i l_i + r(\textbf{v};\lambda) \\
\end{equation}

\begin{equation}\label{equ_spl_original_g}
    \footnotesize
	r(\textbf{v}; \lambda) = - \lambda \sum_{i=1}^N v_i
\end{equation}

\textbf{We believe that determining the difficulty based on loss values is ambiguous, therefore we propose a gradient based method. Gradient based methods always steer towards a certain objective by reducing the error.} Our method dynamically adjusts class weights during training by taking the derivative of loss function w.r.t. class weights instead of using a traditional SP-regularizer. Also, we adjust the segmentation and VQA tasks by taking the derivative of task balance parameter $\alpha$ to steer the training between segmentation and VQA dynamically. These two combined are our main contribution (explained in Section \ref{sec_proposed_CL}). The age parameter $\lambda$ is not used in our method, we use learning rates $\epsilon_{DATAP}$ and $ \epsilon_{DAWEP}$ (explained in detail in Section \ref{sec_proposed_CL}) that adjust the pacing of the curriculum.

Our method is also multimodal since our model works with both image and text data at the same time. In the literature, multimodal curriculum learning is studied in medical report generation \citep{liu_cl_multimodal_medical_report_generation} and speech enhancement \citep{cheng_cl_multimodal_speech_enhancement} but the use CL does not include dynamic adjustment of difficulty with derivatives. Dynamic task prioritization \citep{guo_cl_multitask_dynamic_task_prioritization} is similar to our task adjustment method but key performance metrics (KPIs) that reflect reinforcement learning are preferred over self-paced learning.

\begin{table}[]
\caption{General comparison of the proposed CL method to literature}
\label{table_cl_comparison}
    
    \begin{tabular}{ p{4.0cm} p{4.0cm} p{4.0cm} }
    \hline
    \multicolumn{1}{c}{\textbf{Method}} & \multicolumn{1}{c}{\textbf{Difficulty Measurement}} & \multicolumn{1}{c}{\textbf{Pacing}}  \\ \hline
    Vanilla CL \citep{bengio_cl_entropy}   & Priorly known                               & Entropy of distributions                                   \\
    \vfill Self Paced Learning \citep{kumar_cl_selfpacedlearning} \vfill  & \vfill Measured during training  \vfill  &  \vfill SP-Regularizer                                 \vfill \\ 
    \vfill Self Paced CL  \citep{jiang_self_paced_curriculum_learning} \vfill   &  \vfill Both priorly known and measured during training & \vfill SP-Regularizer                                   \vfill \\ 
    \vfill Teacher-Student CL \citep{hacohen_cl_teacherstudent_initial} \vfill &   \vfill Pretrained teacher model adjusts   & \vfill  Various                                 \vfill \\
    \vfill Implicit CL  \citep{sinha_cl_gaussian_smoothing}  \vfill    &  \vfill Implicitly adjusted by network layers    & Randomly selected batches of data                               \\ 
    \vfill \textbf{Ours}  \vfill   &  \vfill Measured during training with gradient-based adjustment  & Learning rates of gradient descent \\ 
    \end{tabular}
\end{table}

\section{Methodology}\label{sec_methodology}

\subsection{Dataset and Preprocessing}

In this work, we use the FloodNet \citep{rahnemoonfar_floodnet_original} dataset, which contains aerial imagery of the post-disaster damage assessment of Hurricane Harvey, a Category 4 hurricane close to Texas and Louisiana. The data is collected with a Da-Jiang Innovations (DJI) Mavic Pro drone and includes images of Ford Bend County in Texas and other immediately affected locations captured between August 30 and September 04, 2017, with numerous flights. FloodNet has two tracks: semantic segmentation and visual question answering. In the segmentation track, data is annotated with segmentation masks for nine classes that includes building-flooded, building-non-flooded, road-flooded, road-non-flooded, water, tree, vehicle, pool, and grass. In the visual question answering task, images are annotated with different question-answer pairs regarding the details of the image. The objective is to answer different types of questions about a given image. So each image can have more than one question-answer pair, and on average 3.5 question-answer pairs are annotated. Annotated question types are: 'Simple Counting', 'Complex Counting', 'Yes/No', and 'Condition Recognition'. Simple and complex counting questions aim to find the frequency of the object that is present in the image. The difference between simple and complex is that complex counting also considers particular attributes of the object. For example, in simple counting we can ask: "How many buildings are in this image?" whereas, in complex counting, we can ask:" How many buildings are flooded in this image?". Condition recognition questions aim to find the condition of any object present in the image or the condition of the entire image itself as a whole. An example question will be: "What is the condition of the Road?". Finally Yes/No questions aim to find if a particular attribute of an object is true or not. For example: "Is the road flooded?". So FloodNet contains both image and text data, making it a multimodal dataset. The answer classes for the condition recognition questions are: flooded, non-flooded and flooded,non flooded. The answer classes for the Yes/No questions are: just yes or no. Finally the answer classes for the simple and complex counting questions are: integers from 1 to 50 (but not necessarily including all numbers in the given range).

Overall, FloodNet dataset has image data of resolution 4000$\times$3000 in the segmentation track and contains varying numbers of images and annotations for each segmentation class. The order, from highest to lowest, is as follows: tree, road-non-flooded, water, building-non-flooded, vehicle, pool, road-flooded, building-flooded. In the VQA track, The potential distribution of answers varies depending on the type of question. In counting problems, the most common answers are typically '4, 3, 2, 1', while less common answers include '27, 30, 41, 40'. In Condition Recognition problems, the frequent answer is 'non-flooded', and for yes/no questions, the most prevalent response is 'yes'. We can easy say the FloodNet dataset is class imbalanced. However, per class metrics are computed to compensate this (see Table \ref{table_seg_baseline_comparison} and \ref{table_vqa_baseline_comparison}). 

Deep neural networks work with floating point tensor data, so the data available must be preprocessed first in order to be used in training, evaluation and prediction. Preprocessing is done for image and text data with different pipelines. For segmentation, drone images are read from portable network graphics (PNG) files and converted to red, green, and blue (RGB) format. Annotated segmentation masks are integer coded in a single image but separated for each class and converted to grayscale. So each segmentation mask is a separate grayscale image (filled with 255) for each class. Both images and segmentation masks are in unsigned 8-bit integer format, where each pixel value is between 0 and 255. Both images and segmentation masks are resized to be $200\times200$ width and height and normalized by dividing each image and mask by 255. Finally, image and segmentation masks are converted to tensor format to get desired floating-point tensor data. For visual question answering, data is in text format rather than numeric (8-bit integer) like in images and segmentation masks, so preprocessing is different and slightly more complex. Question text needs to be converted to numeric format first. This operation is called tokenization, and there are several methods for doing so. The most common one is word-based tokenization, where each word is split and mapped to a fixed integer. The collection of all word mappings to integers is called "vocabulary". Vocabulary is then stored in high dimensional embedding vectors in floating point tensor format (embeddings are part of the deep neural network model and explained in Section \ref{subsec_dnn}). In this paper, we use character-based embedding, which is the same methodology as word-based embeddings, but individual characters of words (and spaces between words) are tokenized instead of tokenizing words as a whole. This significantly reduces the vocabulary, and eventually the model size. Downside of using character-based tokenization is that token sequences are much longer than word-based tokenization. However, this is not an issue in this work since we only tokenize question sentences, and they are short sequences. There is also a third method: subword-based tokenization \citep{song_fast_wordpiece_tokenization}, where words are broken down into two or more subwords and then tokenized. However, vocabulary size is still relatively large in comparison to character-based tokenization.

So pipeline for preprocessing visual question answering track is as follows: First, question sentences are converted to integer sequences with character-based tokenization, and punctuation characters are removed. A deep neural network model requires that every sequence have the same length, but every question sentence in the dataset might not have the same length. So each sequence is padded with zeros to get the same sequence length. In this way, we can also set the maximum sequence length. Finally, the sequence is converted into the tensor format and ready to be used by the deep neural network.

An example of character-based tokenization is given in Table \ref{tab_ex_char_tokens}. We also added extra tokens to indicate the beginning and end of a word. Start of the sentence is denoted by $<$sos$>$ token, the end of the sentence is denoted by $<$eos$>$ token, the start of a word is denoted by $<$sow$>$ token, and finally, the end of a word is denoted by $<$eow$>$ token.

Answers are also text; however, they are not a sentence and are treated as a fixed number of classes. So each answer is mapped to an integer and converted to probabilities by the loss function (explained in detail in Section \ref{subsub_vqa_evalutaion}). We use PyTorch's cross entropy loss  implementation in our case, so using integers is applicable.

\begin{table}[h]
\centering
\caption{An example question sentence from dataset and corresponding character-based tokenization.}
\label{tab_ex_char_tokens}
\begin{tabular}{l l}
\hline
\textbf{Example Question Sentence} &  \textbf{Character-based Tokenization} \\ \hline
How many objects in total? &  \begin{tabular}[c]{@{}l@{}} $<$sos$>$, $<$sow$>$, h, o, w, $<$eow$>$, \\ $<$sow$>$, m, a, n, y, $<$eow$>$, \\ 
$<$sow$>$,  o, b, j, e, c, t, s, $<$eow$>$,  \\
$<$sow$>$,  i, n, $<$eow$>$, \\
$<$sow$>$,  t, o, t, a, l, $<$eow$>$, $<$eos$>$\\
\end{tabular} \\ 
\end{tabular}
\end{table}

\subsection{Multitask MultiModal Deep Neural Network}\label{subsec_dnn}
\begin{figure}[h]%
\centering
\includegraphics[width=0.9\textwidth]{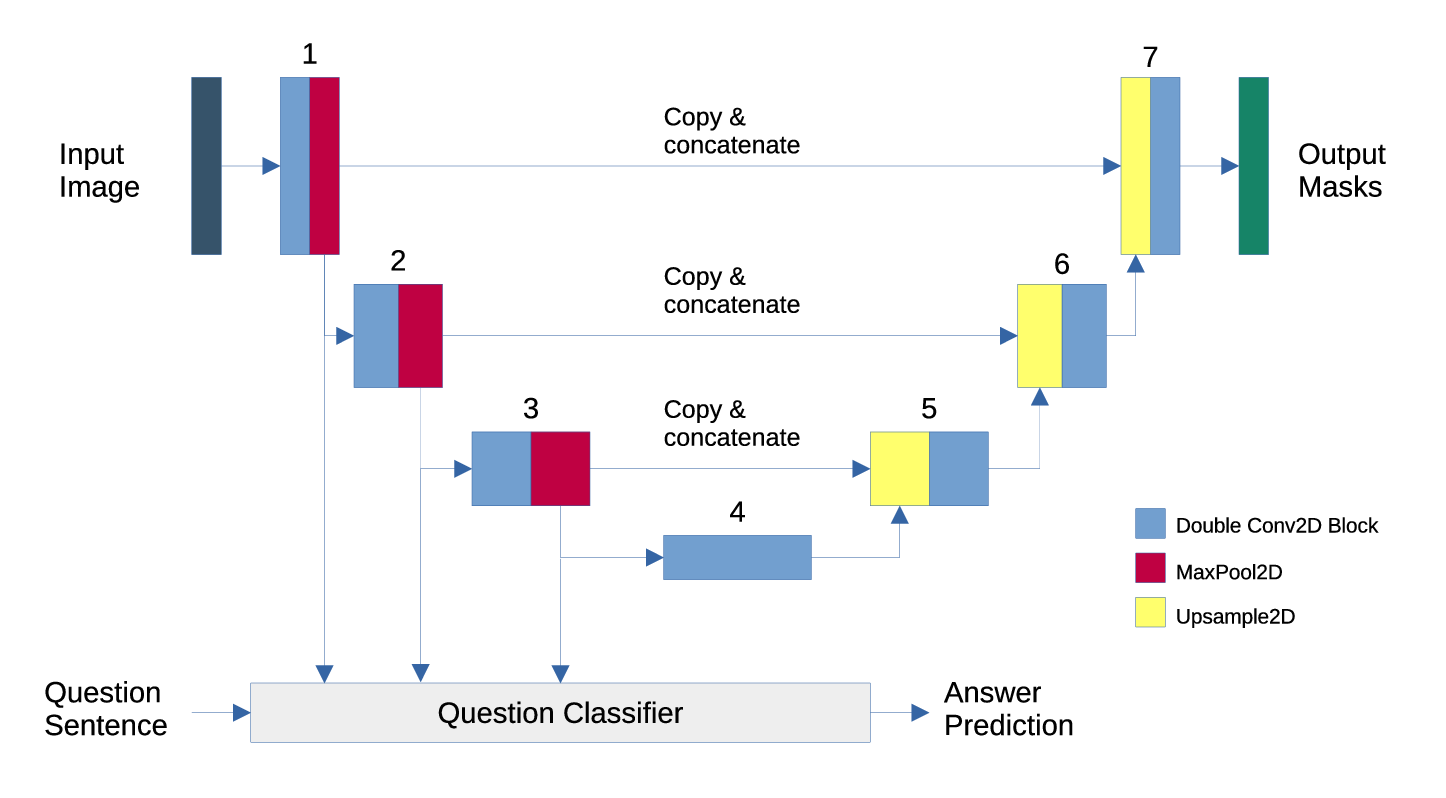}
\caption{Overall structure of our deep neural network. It is a multimodal deep neural network that can process aerial imagery and question sentences in the FloodNet dataset.}\label{fig_multimodal_unet}
\end{figure}

Our deep neural network model architecture is given in Fig. \ref{fig_multimodal_unet}. It can process both image and text (hence it is multimodal) and can also be trained for both segmentation and visual question-answering tasks simultaneously (hence it is multitasking). Image processing is done with U-Net, and question classification is done with our custom question classifier implementation. U-Net is a simple yet effective semantic segmentation model. It is a fully convolutional network (FCN) which means every layer of the convolutional and does not have linear layers like image classifiers have. U-Net resembles a "U" shape and is arranged as downsampling layers followed by upsampling layers with skip connections between them. Each block of U-Net is annotated with numbers (ranging from 1 to 7) in Fig \ref{fig_multimodal_unet}.

The input image, in our case, is an aerial image from FloodNet's segmentation track, and the question is the related question of that image from FloodNet's visual question-answering track. Input images are 200 by 200, 3 channels (red, green, blue), and first downsampled, and features are extracted in blocks 1, 2, and 3. Block 4 is called bottleneck and features are increased without changing any downsampling. In blocks 5, 6 and 7 features are upsampled to predict segmentation masks (in our implementation, each class is a separate segmentation mask). Features from block 1, 2 and 3 are copied and concatenated to upsampling blocks to some of the input image features are passed on without any modification. Since input image is upsampled and downsampled the same way, predicted output segmentation masks are in the same width and height as input images.

Blocks 1, 2, and 3 have double convolutional layers where each has 2D convolution, batch norm, and ReLU activation followed by a 2D max pooling layer. In our implementation, convolutional layers are used only for feature extraction. Convolutional layers have $3\times3$ filters with stride and padding of 1 and don't change the width and height of the input. Downsampling is done in 2D max pooling layers with $2\times2$ filters, reducing feature width and height by half. Block 4 also has double convolutional layers but doesn't downsample the input, hence there is no 2D max pooling. Blocks 4, 5 and 6 first concatenate features passed from blocks 1, 2, and 3, then upsample the features with 2D upsampling layer. In our case, we don't use 2D transposed convolutions, instead, we use upsampling layers to double the feature width and height first and later use double convolution layers to adjust feature maps. Double convolutional layers after upsampling also have $3\times3$ filters with stride and padding of 1 and do not change the width and height of the input. Details about the blocks are given in Table \ref{tab_layer_details}.

\begin{figure}[h]%
\centering
\includegraphics[width=0.9\textwidth]{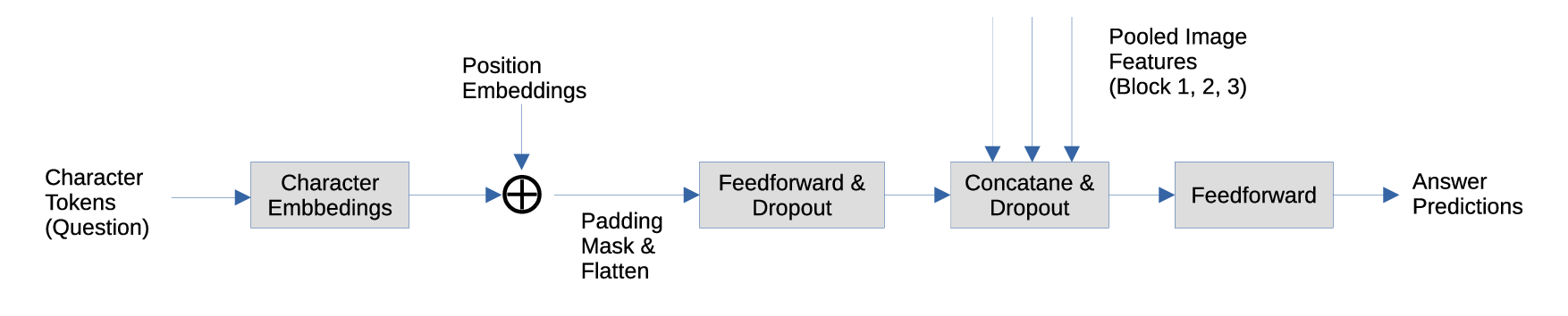}
\caption{Detailed architecture of question classifier mentioned in Fig. \ref{fig_multimodal_unet}. Question classifier takes a question sentence and image features from U-Net and predicts an answer for question with given image.}\label{fig_question_classifier}
\end{figure}

Question classifier implementation details are given in Fig. \ref{fig_question_classifier} and its task is to answer questions that are asked given an image. Questions are tokenized character sequences. Character sequences are represented in trainable embedding vectors. Each character is represented as an 8-dimensional vector in character embeddings. Our model doesn't have any recurrent layers so, in order to preserve sequence's order we add positional embeddings to character embeddings for providing positional information. In this work, positional embeddings are 8-dimensional vectors (the same as character embeddings) that are functions of sine and cosine functions in different frequencies which implement the embeddings used in Transformer architecture \citep{vaswani_vqa_transformer_original}. Even dimensions are generated by $\mbox{PE}_{\mbox{pos},2i} = \mbox{sin}(\mbox{pos}/10000^{2i/d_{emb}})$ and odd dimensions are generated by $\mbox{PE}_{\mbox{pos},2i+1} = \mbox{cos}(\mbox{pos}/10000^{2i/d_{emb}})$ where $pos$ is the position, $i$ is the dimension and $d_{emb}$ is the dimension of the embedding (8 in our implementation). Unlike character embeddings, positional embeddings are not trainable since their sole purpose is to add positional information to the sequence.

Pad tokens is added in preprocessing to ensure the same length for all sequences. However, in practice, pad tokens do not convey any useful information. We use a padding mask to remove their influence in the model by multiplying the embeddings of the positions that correspond to the pad token. The sequence is flatten and passed on to a feedforward linear layer, giving us text features of the question sentence. Image features from block 1 2, and 3 of U-Net are first globally pooled to take the average width and height dimensions and then concatenated with text features. Finally, a feedforward linear layer takes both image and text features and establishes the relation between them to answer to question, given the image.

\begin{table}[]
\centering
\caption{Block details of U-Net (given in Fig. \ref{fig_multimodal_unet}).}
\label{tab_layer_details}

\resizebox{\textwidth}{!}{\begin{tabular}{c c c c }
\hline
Block & \multicolumn{1}{c}{\begin{tabular}[c]{@{}c@{}}Input shape\\ (features$\times$width$\times$height)\end{tabular}} & \multicolumn{1}{c}{\begin{tabular}[c]{@{}c@{}} Output shape \\ (features$\times$width$\times$height)\end{tabular}}  & Layers   \\  \hline 
1     &  3$\times$200$\times$200   & 64$\times$100$\times$100   & \begin{tabular}[c]{@{}c@{}} (Conv2D, BatchNorm2D, ReLU) $\times$ 2  \\ MaxPool2D\end{tabular} \\ \\
2     & 64$\times$100$\times$100    & 128$\times$50$\times$50   & \begin{tabular}[c]{@{}c@{}} (Conv2D, BatchNorm2D, ReLU) $\times$ 2  \\ MaxPool2D\end{tabular} \\ \\
3     & 128$\times$50$\times$50    & 256$\times$25$\times$25  & \begin{tabular}[c]{@{}c@{}} (Conv2D, BatchNorm2D, ReLU) $\times$ 2  \\ MaxPool2D\end{tabular} \\ \\
4     &  256$\times$25$\times$25  & 512$\times$25$\times$25  & \begin{tabular}[c]{@{}c@{}} (Conv2D, BatchNorm2D, ReLU) $\times$ 2\end{tabular}  \\ \\
5     & 768$\times$25$\times$25   & 256$\times$50$\times$50  & \begin{tabular}[c]{@{}c@{}} Upsample2D \\ (Conv2D, BatchNorm2D, ReLU) $\times$ 2\end{tabular} \\ \\
6     & 384$\times$50$\times$50  & 256$\times$100$\times$100    & \begin{tabular}[c]{@{}c@{}} Upsample2D \\ (Conv2D, BatchNorm2D, ReLU) $\times$ 2\end{tabular}  \\ \\
7     & 320$\times$100$\times$100   & 128$\times$200$\times$200  & \begin{tabular}[c]{@{}c@{}} Upsample2D \\ (Conv2D, BatchNorm2D, ReLU) $\times$ 2\end{tabular} \\ \\
\end{tabular}}
\end{table}

\subsection{Losses and Evaluation Metrics}

\subsubsection{Segmentation Loss and Evaluation}

Segmentation output is an image that contains masks for different classes. It is formulated as a pixel-level binary classification task. So loss function that is used for segmentation is called the binary cross-entropy (BCE) function. Binary cross-entropy function is given in Eq. \ref{eq_bce} where $N$ is the number of pixels in current batch, $p$ is the target pixel value and $\hat{p}$ is the predicted pixel value by our model. Since predicted and target segmentation masks are 2-dimensional pixels for each class, masks are flattened to have a single dimension of pixels. By flattening the masks, the pixels are arranged in a linear sequence, making it possible to compare them as if they were binary class probabilities. In other words, each pixel in the flattened mask can be treated as a value that represents the likelihood or probability of belonging to a particular class.

\begin{equation}\label{eq_bce}
\mathcal{L}_{SEG}(p,\hat{p}) = - \frac{1}{N} \sum_{i=1}^N p_i\log(\sigma(\hat{p_i})) + (1-p_i)\log(1-\sigma(\hat{p_i}))
\end{equation}

For Eq. \ref{eq_bce} to work, predicted pixel values ($\hat{p}$) must be between 0 and 1. To achieve this, output masks (final layer of U-Net) is passed on a sigmoid function (denoted with $\sigma$) to ensure that pixel values are between 0.0 and 1.0. 

For the evaluation of segmentation task, we use mean intersection over union (mIoU), which is a performance metric used to evaluate the accuracy of segmentation tasks, such as identifying objects or regions within an image. This metric is given in Eq. \ref{eq_mIoU} where $N$, is the number of segmentation masks, $P$ is the pixel map of target segmentation masks and $\hat{P}$ is the pixel map of predicted segmentation masks. In Eq. \ref{eq_mIoU}, the overlap between predicted masks and target masks is computed. The numerator is called the intersection and it is the area of overlap between the target and predicted masks. The denominator is called the union and it is the the area covered by both the predicted and target masks. In order to correctly compute intersection and union, output predictions are passed into a sigmoid function to squeeze values between 0.0 and 1.0. Then, pixel values over the threshold of 0.5 are considered as 1, and others are considered as 0. Finally, we take the mean of intersection over union.

\begin{equation}\label{eq_mIoU}
mIoU(P, \hat{P}) = \frac{1}{N}\sum_{i=1}^N \frac{\vert P_i \cap \hat{P_i} \vert}{\vert P_i \cup \hat{P_i} \vert} =  \frac{1}{N}\sum_{i=1}^N  \frac{\vert P_i \cap \hat{P_i} \vert}{\vert P_i \vert + \vert \hat{P_i} \vert - \vert P_i \cap \hat{P_i} \vert}
\end{equation}

Mean intersection over union can also be defined in Eq. \ref{eq_mIoU_baseline_form} where the intersection between predicted and target masks are denoted by true positives (TP) and union is denoted by the summation of true positives (TP), false positives (FP) and false negatives (FN) respectively.

\begin{equation}\label{eq_mIoU_baseline_form}
mIoU = \frac{1}{N}\sum_{i=1}^N \frac{TP_i}{TP_i + FP_i + FN_i}
\end{equation}

\subsubsection{Visual Question Answering Loss and Evaluation}\label{subsub_vqa_evalutaion}
In visual question answering, each answer is a fixed class. Therefore, evaluation is formulated as multi-class classification task. Negative log-likelihood loss for multi-class classification is formulated as individual losses given in Eq. \ref{equ_vqa_individual} for $N$ answer predictions in the current batch. 


\begin{equation}\label{equ_vqa_individual}
\mathcal{L}_{VQA}(x,y) = \{ \ell_1, \ell_2, ..., \ell_N \}
\end{equation}

Each answer prediction is softmax activated and weighted by corresponding class weight $\omega_{y_n}$ as given in the Eq. \ref{equ_vqa_weighted_each}. 

\begin{equation}\label{equ_vqa_weighted_each}
\ell_n = -\omega_{y_n}\log\frac{e^{x_{n,y_n}}}{\sum_{c=1}^C e^{x_{n,c}}} \ \ \ for\ n=1,2,\dots,N
\end{equation}

Loss is normalized by sum of all class weighted in current batch as given in Eq. \ref{equ_vqa_weighted}.

\begin{equation}\label{equ_vqa_weighted}
\mathcal{L}_{VQA}(x,y) = \frac{1}{\sum_{n=1}^N \omega_{y_n}} \sum_{n=1}^N \ell_n
\end{equation}

Final loss function for visual question answering ($\mathcal{L}_{VQA}$) is given in Eq. \ref{equ_vqa_final}.

\begin{equation}\label{equ_vqa_final}
\mathcal{L}_{VQA}(x,y) = \frac{1}{\sum_{n=1}^N \omega_{y_n}} \sum_{n=1}^N (-\omega_{y_n}\log\frac{e^{x_{n,y_n}}}{\sum_{c=1}^C e^{x_{n,c}}}) \ \ \ for\ n=1,2,\dots,N 
\end{equation}

The evaluation metric for visual question answering is accuracy. It is calculated using the equation shown in Eq. \ref{eq_vqa_acc_baseline_form}, where the batch size of the current data is represented by $N$, and true positives, true negatives, false positives, and false negatives are denoted as TP, TN, FP, and FN, respectively.

\begin{equation}\label{eq_vqa_acc_baseline_form}
Accuracy_{VQA} =  \frac{1}{N}\sum_{i=1}^N \frac{TP_i + TN_i}{TP_i + TN_i + FP_i + FN_i}
\end{equation}

\subsubsection{Total Loss}
Putting two tasks together, we get the total loss equation given in Eq. \ref{equ_total_loss} with task balance parameter alpha, given as $\alpha$. The parameter $\alpha$ controls the emphasis between segmentation and visual question answering tasks. The value of $\alpha$ should be between value 0.0 and 1.0, setting $\alpha$ to 0.5 means that both tasks are learned in balanced manner. Setting $\alpha$ to 0.0 means that model solely focus on visual question answering while ignoring segmentation. Conversely, setting $\alpha$ to 1.0 means that model solely focus on segmentation while ignoring visual question answering. This work implements a dynamic $\alpha$ adjustment method (details are explained in Section \ref{subsubsec_datap}). Note that the total loss function is multi-task and multimodal since visual question answering and segmentation are learned simultaneously and both image and text data are used.

\begin{equation}\label{equ_total_loss}
\mathcal{L}_{Total} = \alpha \mathcal{L}_{VQA} + (1-\alpha) \mathcal{L}_{SEG}
\end{equation}

\subsection{Proposed Curriculum Learning Method}\label{sec_proposed_CL}
In various tasks, curriculum learning techniques have been successfully applied across the board in machine learning (see Section \ref{subsec_lit_review_cl}). However, the use of the curriculum techniques may be constrained by the requirement to establish a method of classifying the samples from easy to hard and the proper pacing function for introducing more difficult data. \textbf{In this paper, we take an alternative approach to self-paced learning where instead of measuring difficulty explicitly or knowing it prior, we implement a gradient-based method that can automatically adjust the difficulty during the training.} With the proposed method, we let the model choose for itself which tasks and classes it will focus on. Therefore, the model will decide its curriculum by adjusting the data to training next.

Our model is multimodal and contains segmentation and visual question answering tasks at the same time. So, our objective was to develop a curriculum learning approach to balance the tasks and adjust the difficulty at the same time. Following the main idea of curriculum learning, from easy to hard examples, our main contribution is explained in Sections \ref{subsubsec_datap}, \ref{subsubsec_DAWEP}, and \ref{subsubsec_DATWEP}, given below.

\subsubsection{Dynamic Task Prioritization Curriculum Learning}\label{subsubsec_datap}

Dynamic Task Prioritization (DATAP) Curriculum Learning is a method for dynamically setting the task balancing parameter $\alpha$ during model training. The main intuition behind DATAP is for the model to make its own decisions about how much priority it should give between two tasks. It does not rely on predefined functions of difficulty or pacing that might be based on SPL. It is achieved by taking the derivative of the total loss $\mathcal{L}_{Total}$ w.r.t $\alpha$ given in Eq. \ref{equ_datap_basic} and updating the $\alpha$ value with gradient descent given in Eq. \ref{equ_datap_gradient_descent} with a learning rate of $\epsilon_{DATAP}$.

\begin{equation}\label{equ_datap_basic}
\frac{\partial \mathcal{L}_{Total}}{\partial \alpha} = \mathcal{L}_{VQA} - \mathcal{L}_{SEG}
\end{equation}

\begin{equation}\label{equ_datap_gradient_descent}
\alpha = \alpha - \epsilon_{DATAP} \left( \frac{\partial \mathcal{L}_{Total}}{\partial \alpha} \right)
\end{equation}

However, the solution given in Eq. \ref{equ_datap_gradient_descent} is not sufficient because $\alpha$ is also required to be in the interval [0.0, 1.0]. Otherwise, training goes unstable, and the overall performance of the model is significantly reduced. For this reason, a regularization loss is $\mathcal{L}_{\alpha_{reg}}$ defined in Eq. \ref{equ_datap_regularization_loss} which is the absolute value of $\alpha$ subtracted from sigmoid of $\alpha$. This absolute value will be very small when $\alpha$ is within the interval of [0.0, 1.0] but as $\alpha$ shoots off from the range [0.0, 1.0], regularization will increase, effectively keeping $\alpha$ values between the intended interval of [0.0, 1.0].


\begin{equation}\label{equ_datap_regularization_loss}
\mathcal{L}_{\alpha_{reg}} = \lvert \alpha - \sigma(\alpha)  \rvert, \ \ \ \ \sigma(\alpha) = \frac{1}{1 + e^\alpha}
\end{equation}

Then, a regularization term is added to gradient descent in Eq. \ref{equ_datap_gradient_descent} and we get Eq. \ref{equ_datap_with_reg_term}. By default, the regularization term is very strict and can keep $\alpha$ values strictly around 0.5 and this can cause performance degradation. For this reason, an additional scalar $\lambda_{\alpha_{reg}}$ term that is multiplied by the regularization term to lower regularization impact. The value of $\lambda_{\alpha_{reg}}$ term is chosen as 0.75 and doesn't change during the training. Given the maximum and minimum regularization such that $0.5 \pm 0.5 \times 0.75 = [0.12, 0.87]$, the $\lambda_{\alpha_{reg}}$ value of 0.75 gives us a nice interval of $\alpha$ values.

\begin{equation}\label{equ_datap_with_reg_term}
\alpha = \alpha - \epsilon_{DATAP}(\frac{\partial \mathcal{L}_{Total}}{\partial \alpha}  + \lambda_{\alpha_{reg}} \frac{\partial \mathcal{L}_{\alpha_{reg}}}{\partial \alpha})
\end{equation}

Putting everything together, the final formula of updating $\alpha$ with gradient descent is given in Eq. \ref{equ_datap_final}. Since updates are regularized, the chance of gradient vanishing and explosion is small and there is no need for gradient clipping. Derivation of Eq. \ref{equ_datap_final} involves several steps and is explained in \ref{app_datap_derivation}.

\begin{equation}\label{equ_datap_final}
\alpha = \alpha - \epsilon_{DATAP}((\mathcal{L}_{VQA} - \mathcal{L}_{SEG}) + \lambda_{\alpha_{reg}} (\frac{\alpha-\sigma(\alpha)}{\vert \alpha-\sigma(\alpha) \vert} (1-\sigma(\alpha)+\sigma^2(\alpha)))
\end{equation}

\subsubsection{Dynamic Weight Prioritization Curriculum Learning}\label{subsubsec_DAWEP}
Like DATAP, Dynamic Weight Prioritization (DAWEP) is a method for dynamically adjusting each class weight of all available classes during model training. It is only applied to VQA answer classes and achieved by taking the derivative of the VQA loss $\mathcal{L}_{VQA}$ w.r.t each class weight $w_n$ given in Eq. \ref{equ_vqa_final} and updating the $w_n$ value with gradient descent given in Eq. \ref{equ_DAWEP_descent}. For weights $w_i$ for $i \in \{1,\, \ldots,\,N\}$ where each $w_n$ is a class weight and $N$ is total number of classes, gradients are defined in Eq. \ref{equ_DAWEP_gradients}.

\begin{align}\label{equ_DAWEP_gradients}
   \nabla_{\mathbf{w}}\mathcal{L}_{VQA} = \begin{bmatrix}
           \frac{\partial \mathcal{L}_{VQA}}{\partial w_1} \\[6pt]
           \frac{\partial \mathcal{L}_{VQA}}{\partial w_2} \\[6pt]
           \frac{\partial \mathcal{L}_{VQA}}{\partial w_3} \\[2pt]
           \vdots \\
           \frac{\partial \mathcal{L}_{VQA}}{\partial w_n}
        \end{bmatrix}
\end{align}

Calculating gradients for each class in Eq. \ref{equ_DAWEP_gradients} yields the Eq. \ref{equ_DAWEP_derivative} where $\hat{n}$ is the selected samples that belong to the class $n$ among all samples in that batch, $\hat{n}_{occurance}$ is the total number of occurrences of class $\hat{n}$ in current batch, $x_{y_{\hat{n}}}$, $y_{\hat{n}}$, $\hat{C}$ are the predictions, targets and total number of classes for the the selected class respectively. Derivation of Eq. \ref{equ_DAWEP_derivative} is explained in \ref{app_dawep_derivation}.

\begin{align}\label{equ_DAWEP_derivative}
\frac{\partial \mathcal{L}_{VQA}}{\partial w_{\hat{n}}} = \left( -\frac{\hat{n}_{occurance}}{(\sum_{n=1}^N \omega_{y_n})^2}
\sum_{n=1}^N \omega_{y_n}\log\frac{e^{x_{n,y_n}}}{\sum_{c=1}^C e^{x_{n,c}}} \right) + \nonumber \\
\left(  \frac{1}{\sum_{n=1}^N \omega_{y_n}}(\log\frac{e^{x_{y_{\hat{n}}}}}{\sum_{c=1}^{\hat{C}} e^{x_{n,c}}}) \right)
\end{align}

After taking the derivative, all weights are updated using gradient descent given in Eq. \ref{equ_DAWEP_descent} with learning rate $\epsilon_{DAWEP}$. Gradients are clipped in the range of [-1.5, 1.5] so that gradient explosion is prevented ($\left[ \cdots \right]_{[-1.5,1.5]}$ denotes clipping of gradient value for a maximum of 1.5 and minimum of -1.5). 

\begin{equation}\label{equ_DAWEP_descent}
w_n = w_n - \epsilon_{DAWEP} \left( \left[ \frac{\partial \mathcal{L}_{VQA}}{\partial w_n} \right]_{[-1.5,1.5]} \right) \ \ \ for\ n=1,2,\dots,N 
\end{equation}

The main intuition behind DATWEP is for the model to choose the best weight for each class given the VQA loss, so the emphasis on which classes to focus on should be selected by the model itself, not by any common sense CL difficulty or pacing functions. In this work, DATWEP is limited to VQA however, theoretically it can be applied to any classification task.

\subsubsection{Dynamic Task and Weight Prioritization Curriculum Learning}\label{subsubsec_DATWEP}

Combining both approaches given in sections \ref{subsubsec_datap} and \ref{subsubsec_DAWEP}, we get Dynamic Task and Weight Prioritization (DATWEP) Curriculum Learning given in Algorithm \ref{DATWEP_cl_alg}. 
 
\begin{align*}
\alpha = \alpha - \epsilon_{DATAP} \left( \frac{\partial \mathcal{L}_{Total}}{\partial \alpha}  + \lambda_{\alpha_{reg}} \frac{\partial \mathcal{L}_{\alpha_{reg}}}{\partial \alpha} \right) \\ \\
w_n = w_n - \epsilon_{DAWEP} \left( \left[ \frac{\partial \mathcal{L}_{VQA}}{\partial w_n} \right]_{[-1.5,1.5]} \right)
\end{align*}

\begin{algorithm}[!h]
\caption{DATWEP CL Algorithm}
\label{DATWEP_cl_alg}
\begin{algorithmic}[1]
 \renewcommand{\algorithmicrequire}{\textbf{}}
 \Require $M$ -- Multimodal deep learning model;
 \Require $\mathcal{D}=\{x^{SEG}_i, x^{VQA}_i, y^{SEG}_i, y^{VQA}_i,\}$ -- training data set with images and text;
 \Require $P=\{p^{SEG}_i, p^{VQA}_i,\}$  -- model predictions for segmentation and VQA;
 \Require $E$ -- number of iterations (epochs);
 \Require $\mathcal{L}^{SEG}$ -- segmentation loss function (given by Eq. \ref{eq_bce});
 \Require $\mathcal{L}^{VQA}$ -- VQA loss function (given by Eq. \ref{equ_vqa_final});
 \Require $\alpha$ -- task balance parameter;
 \Require $\mathcal{W}=\{w_1, w_2, \dots, w_N \}$ -- class weights (for total $N$ classes); 
 
 \State Initialize $\alpha = 0.5$, set all class weights in $\mathcal{W}$ to 1.0
  \For{$i = {1,2,...,E}$}
    \State $p^{SEG}, p^{VQA} \leftarrow M(x^{SEG}, x^{VQA})$
    \State $\mathcal{L}^{Total} \leftarrow \alpha * \mathcal{L}^{SEG}(p^{SEG}, y^{SEG}) + (1 - \alpha) * \mathcal{L}^{VQA}(p^{VQA}, y^{VQA})$
    \State $M \leftarrow train(M)$ 
    \State Update $\alpha$ using Eq. \ref{equ_datap_final}
    \State Update class weights $\mathcal{W}$ using Eq. \ref{equ_DAWEP_descent}
   \EndFor
\end{algorithmic}
\end{algorithm}

\section{Experiments}\label{sec_experiments}


The FloodNet dataset is split into training, validation, and test, sets with 70\% for training, 15\% for validation and 15\% for testing. We used AdamW optimizer for faster convergence with a learning rate of 0.001. The learning rate is scheduled during the training with a step scheduler and reduces the learning rate by 95\% at every three epochs. The learning rate for DATAP ($\epsilon_{DATAP}$) is selected as 0.002 and for DATWEP ($\epsilon_{DAWEP}$) selected as 0.001 and also gradients are clipped (between [-1.5, 1.5]) to prevent exploding gradients problems during training. Training is done for 25 epochs in total. Training algorithm with supervised training of the model with DATWEP is given in Algorithm \ref{DATWEP_cl_alg}. For the VQA task, the maximum question length is chosen as 70 characters in total. Length of 70 characters is chosen by looping through all possible question sentences in the dataset and finding the maximum question length. This way we can include every question sentence without any truncation. Model implementation and all experiments are conducted with PyTorch 2.0 framework.

\subsection{Results and Discussion}

\begin{figure}[h]%
\centering
\includegraphics[width=\textwidth,height=\textheight,keepaspectratio]{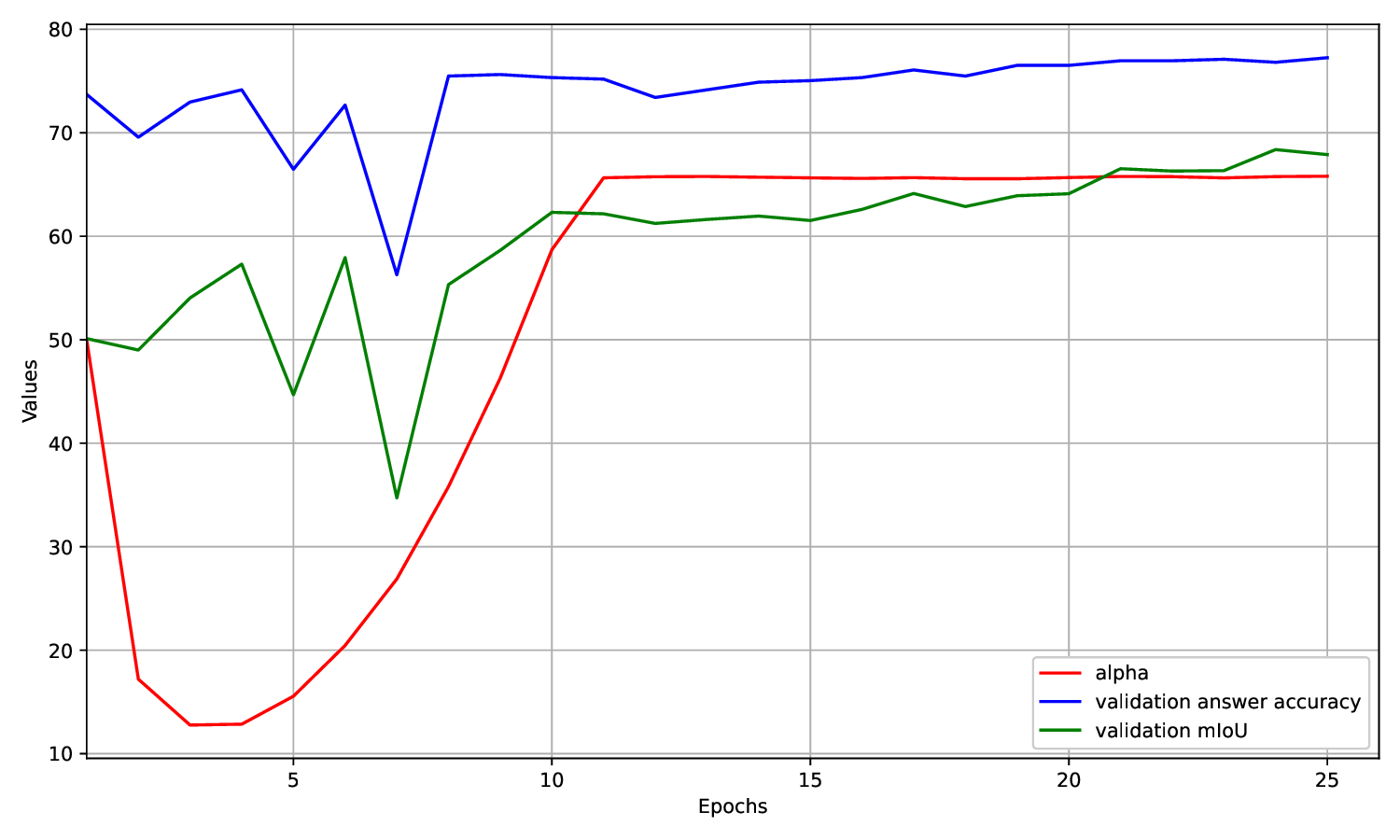}
\caption{Validation metrics for both VQA and segmentation with alpha values over Epochs. Note that alpha values normally within the interval of [0.0, 1.0] but here multiplied by 100 to compare with other metrics.}\label{fig_alpha_with_metrics}    
\end{figure}

In this section validation metrics along with the change of task prioratization parameter $\alpha$ (alpha) are given in Fig. \ref{fig_alpha_with_metrics}, performance comparison of our model and method with FloodNet baselines \citep{rahnemoonfar_floodnet_original} for semantic segmentation is given in Fig. \ref{table_seg_baseline_comparison} and for VQA is in Fig.  \ref{table_vqa_baseline_comparison}. Example predictions are given in \ref{sec_app_example_predictions_multimodal} and \ref{sec_app_example_predictions_vqa_only}, where an example of actual and predicted segmentation masks are given in Fig. \ref{fig_masks_actual} and Fig. \ref{fig_masks_prediction} and finally an example of VQA answer prediction example is given in Fig. \ref{fig_vqa_pair_1}.

The value of $\alpha$ and its change over training time is given in Fig. \ref{fig_alpha_with_metrics}. It starts at its initial value of 0.5, so training starts with equal emphasis on semantic segmentation and VQA tasks but shifts to VQA first, then shifts back to segmentation later. Investigating the Fig. \ref{fig_alpha_with_metrics}, first, the DATAP decided to focus on VQA by lowering $\alpha$ value since its accuracy is higher, indicating that it is an easier task than semantic segmentation for the start. But over time, VQA performance saturates (around epoch 3), and the DATAP increases $\alpha$ value to focus training to semantic segmentation. Eventually, both tasks are saturated and $\alpha$ is stabilized to keep a stable performance among the VQA and semantic segmentation tasks.

Initial and final VQA class weights are in Table \ref{table_intial_and_final_class_weights}. Initial weights for all VQA classes are set to 1.0 uniformly. This uniform initialization assumes an equal level of difficulty for all classes, providing a baseline for the model to start learning. By treating all classes equally at the beginning, the model can initially focus on understanding the basic patterns and concepts shared across the dataset. Investigating these results, we can observe that DAWEP decided that model should focus mostly on Non-Flooded cases and least on counting cases. This indicates that the counting questions are the most challenging questions. Yes/No classes have the same impact and their difficulty is slightly harder than Non-Flooded classes. Finally, we observe that Flooded and Flooded, Non-Flooded classes are almost unchanged and have little effect on the difficulty of learning.

Counting answers in VQA can be more challenging than condition recognition or yes/no answers due to several reasons. Counting involves dealing with ambiguity, variability, and context in visual data, making it difficult to accurately identify and count objects. It requires a deep understanding of relationships, spatial arrangements, and estimation. Additionally, limited training data with accurately annotated counts can hinder model generalization. In contrast, condition recognition or yes/no answers focus on specific attributes or conditions, which are comparatively easier to identify and require less estimation. This is also noted in the literate. \cite{antol_vqa_original} showed that counting answering has the accuracy of 36.77\% while yes/no answers has the accuracy of 80.50\% even with the best models. \cite{lobry_rsvqa_original} showed that counting answers has the lowest accuracy of 68.63\% among presence, comparison and area answers that have the accuracy range between 76.33\% - 90.43\% using RSVQA dataset (a very similar dataset to FloodNet). \cite{yuan_vqa_easytohard_cl_remote_sensing} applied the curriculum learning to RSVQA and increased the counting answer accuracy to 69.22\% but it is still the lowest. Overall we can say the counting questions are the most challenging. Therefore, we believe the final answer weights that we found are consistent with the similar studies.

%
%

A comparison of our results with segmentation baselines is shown in Table \ref{table_seg_baseline_comparison}. Since DATWEP is only applied to VQA tasks, we expect no improvement over baselines. However, in Table \ref{table_vqa_baseline_comparison}, we see that DATWEP improved performance comparing other VQA baselines. This is due to the fact that DATWEP selects the best $\alpha$ value and class weights for the training and improves the VQA performance.

\begin{table}[H]
\centering
\caption{Initial and Final (End of Training) Weights for Every VQA Answer Class. Weights are Adjusted During Training by DAWEP Method.}
\label{table_intial_and_final_class_weights}

\resizebox{9cm}{!}{	
    \begin{tabular}{l c c}
    \hline
    \multicolumn{1}{c}{\textbf{Answer Class}}                             & \textbf{Initial Weight} & \textbf{Final Weight} \\ \hline
    Flooded                                                                 & 1.0                     & 1.0039              \\ 
    Non-Flooded                                                             & 1.0                     & 2.2706                \\ 
    Flooded,Non-Flooded                                                     & 1.0                     & 0.9515               \\ 
    Yes                                                                     & 1.0                     & 1.2639                \\ 
    No                                                                      & 1.0                     & 1.2700               \\ 
    \begin{tabular}[c]{@{}l@{}}Counting (average)\end{tabular}              & 1.0                     & 0.9175                  \\ 
    \end{tabular}
}
\end{table}

\begin{table}[H]
	\centering
	\caption{Mean Intersection over Union (mIoU) Comparison with FloodNet Baselines for Semantic Segmentation on Testing Set}\label{table_seg_baseline_comparison}
    \resizebox{\textwidth}{!}{	
	\begin{tabular}{l c c c c c c c c c c}
	\hline
		Method & Building Flooded & \begin{tabular}{@{}c@{}}Building Non \\  Flooded\end{tabular} & Road Flooded &  \begin{tabular}{@{}c@{}}Road Non \\  Flooded \end{tabular} & Water & Tree & Vehicle & Pool & Grass & mIoU\\ 
		\hline
		ENet \citep{paszke_enet} & 21.82 & 41.41 & 14.76 & 52.53 & 47.14 & 62.56 & 26.21 & 16.57 & 75.57 & 39.84 \\ 
		DeepLabV3+ \citep{chen_encoder_decoder_sep_segmentation} & 28.10 & 78.10 & 32.00 & 81.10 & 73.00 & 74.50 & 33.60 & 40.00 & 87.10 & 58.61 \\ 
  
		\textbf{U-Net with VQA (Ours)} & 47.87 & 54.76 & 38.94 & 55.19 & 58.86 & 72.30 & 18.83 & 34.55 & 79.26 & 67.50 \\ 
  
		PSPNet \citep{zhao_pyramid_scene_parsing} & 65.61 & 90.92 & 78.69 & 90.90 & 91.25 & 89.17 & 54.83 & 66.37 & 95.45 & 80.35 \\ 
	\end{tabular}

	}
	
\end{table}

\begin{table}[H]
    \centering
    \caption{Accuracy Comparison with FloodNet Baselines for VQA}\label{table_vqa_baseline_comparison}
    \vspace{.2cm}

    \resizebox{\textwidth}{!}{

    \begin{tabular}{cc c c c c c}
        \\ \hline
        \multicolumn{1}{c}{\textbf{Method}}
        & \multicolumn{1}{c}{\textbf{Data Type}} & \textbf{Overall Accuracy} &
        \textbf{\begin{tabular}[c]{@{}c@{}}Accuracy for \\ 'Simple
                Counting'\end{tabular}} & \textbf{\begin{tabular}[c]{@{}c@{}}Accuracy for \\
                'Complex Counting'\end{tabular}} & \textbf{\begin{tabular}[c]{@{}c@{}}Accuracy
                for \\ 'Yes/No'\end{tabular}} & 
        \textbf{\begin{tabular}[c]{@{}c@{}}Accuracy for \\ "Condition\\
                Recognition"\end{tabular}} \\ \hline

        \multicolumn{1}{c}{\multirow{2}{*}{Concatenation \citep{zhou_simple_baseline_vqa}}} 	       
        & \multicolumn{1}{c}{Validation}
        & 0.53 		     
        & 0.06
        & 0.05
        & 0.33
        & 0.88
     
        \\ 
        \multicolumn{1}{c}{}
        & \multicolumn{1}{c}{Testing}		 
        & 0.52 		   
        & 0.06
        & 0.03    
        & 0.31
        & 0.88
        
        \\ 
        \multicolumn{1}{c}{\multirow{2}{*}{Point-wise Multiplication
                 \citep{antol_vqa_original}}} 
        & \multicolumn{1}{c}{Validation}	 
        & 0.77
        & 0.32
        & 0.29
        & 0.95
        & 0.96
        
        \\ 
        \multicolumn{1}{c}{}
        & \multicolumn{1}{c}{Testing}		 
        & 0.77		    
        & 0.3
        & 0.28
        & 0.97
        & 0.97
        
        \\ 
        \multicolumn{1}{c}{\multirow{2}{*}{SAN \citep{yang_stacked_attention_vqa}}}
        & \multicolumn{1}{c}{Validation} 	
        & 0.7
        & 0.31
        & 0.30
        & 0.58
        & 0.97
        
        \\ 
        \multicolumn{1}{c}{}
        & \multicolumn{1}{c}{Testing}		 
        & 0.71 		     
        & 0.32
        & 0.32
        & 0.62
        & 0.96
        
        \\ 
        \multicolumn{1}{c}{\multirow{2}{*}{MFB with Co-Attention \citep{yu_factorized_vqa}}}		       
        & \multicolumn{1}{c}{Validation}
        & 0.77 		     
        & 0.27
        & 0.26
        & 0.98
        & 0.97
        
        \\ 
        \multicolumn{1}{c}{}
        & \multicolumn{1}{c}{Testing}		 
        & 0.74    
        & 0.21
        & 0.21
        & 0.98
        & 0.96
        
        \\ 
        
        \multicolumn{1}{c}{\multirow{2}{*}{\textbf{DATWEP (Ours)}}}		       
        & \multicolumn{1}{c}{Validation}
        & 0.77 
        & 0.23 
        & 0.32 
        & 0.98 
        & \textbf{0.98}  
        
        \\ 
        \multicolumn{1}{c}{}
        & \multicolumn{1}{c}{Testing}		 
        & \textbf{0.78}	
        & 0.27 
        & 0.28 
        & \textbf{0.99} 
        & \textbf{0.98} 
        
        \\ 
        
    \end{tabular}

    }

\end{table}

\section{Conclusions and Future Work}\label{sec_conclusions_futurework}
In this work, we studied post-disaster analytics with the FloodNet dataset. FloodNet has two tracks: semantic segmentation and VQA, and has annotations for ten semantic segmentation classes and 41 different answer classes for VQA. We build a multi-task multimodal model that can be trained in both tasks simultaneously with both image and text data. Due to the challenging nature of multimodal training, two curriculum learning methods are proposed: DATAP and DAWEP. Putting two methods together, DATWEP is used as the main contribution and training methodology for this paper. DATAP is applied to the task balancing $\alpha$, and DAWEP is applied to weights of VQA classes. Performance comparisons are made with other FloodNet baselines for both semantic segmentation and VQA, and it can be seen that VQA performance is increased (especially for counting problems which are the hardest). We found that task balancing parameter $\alpha$ favors the task with the highest performance metric first and then moves to another task until performance is saturated for both tasks. Also, we observe that the Non-Flooded answers are the most, and the counting answers are the least prioritized. These observations implies that the model itself can automatically adjust difficulty and pacing. Finally, results with DATWEP are compared with other FloodNet baseline results, and we see that DATWEP shows performance improvement for the VQA task.

In future studies, we would like to use we would like to apply DATAP, DAWEP, and DATWEP to other deep learning tasks and datasets. One important point to consider is understanding how weight balance for an imbalanced dataset will perform on DAWEP. We believe curriculum learning has potential for solving many performance issues in deep learning in general.


\section{Statement of Funding}
This research did not receive any specific grant from funding agencies in the public, commercial, or not-for-profit sectors.

\section{Declaration of Generative AI and AI-assisted technologies in the writing process}
During the preparation of this work the author(s) rarely used ChatGPT in order to improve the grammar of few sentences.  After using this tool/service, the author(s) reviewed and edited the content as needed and take(s) full responsibility for the content of the publication.

\appendix
\section{Derivation of DATAP and DAWEP}
\subsection{Derivation of DATAP with Regularization}\label{app_datap_derivation}

Derivation of DATAP without regularization is straightforward, taking the derivate of total loss (given in Eq. \ref{equ_total_loss}) w.r.t to $\alpha$ will give us Eq. \ref{equ_datap_basic}. Derivation of DATAP regularization on the other hand, requires more steps given as below: 

\bigskip

\begin{equation}\label{equ_app_datap_regularization_loss}
\frac{\partial \mathcal{L}_{\alpha_{reg}}}{\partial \alpha} = \frac{\partial (\lvert \alpha - \sigma(\alpha)  \rvert)}{\partial \alpha}, \ \ \ where \ \ \sigma(\alpha) = \frac{1}{1 + e^\alpha}
\end{equation}

\bigskip

Using the derivate rule of absolute functions given in Eq. \ref{equ_app_derivative_of_absolute} where $f(x)$ is any differentiable function:

\begin{equation}\label{equ_app_derivative_of_absolute}
\lvert f(x) \rvert ' = \frac{f(x)}{\lvert f(x) \rvert} f'(x)
\end{equation}

\bigskip

Substituting $\mathcal{L}_{\alpha_{reg}}$ into $f(x)$ we get Eq. \ref{equ_app_derivative_of_absolute_subs}:

\bigskip

\begin{equation}\label{equ_app_derivative_of_absolute_subs}
\frac{\partial \mathcal{L}_{\alpha_{reg}}}{\partial \alpha} = \frac{\alpha-\sigma(\alpha)}{\vert \alpha-\sigma(\alpha) \vert} (1-\sigma'(\alpha))
\end{equation}

\bigskip

As can be seen in Eq. \ref{equ_app_derivative_of_absolute_subs}, derivative of sigmoid function should be computed as well. Using the derivate of sigmoid given in Eq. \ref{equ_app_derivative_of_sigmoid}, we get the final form of DATAP regularization in Eq \ref{equ_app_datap_regularization_loss_derivative}:

\begin{equation}\label{equ_app_derivative_of_sigmoid}
\frac{\partial \sigma(\alpha)}{\partial \alpha} = \sigma(\alpha)(1-\sigma(\alpha)) = \sigma(\alpha) - \sigma^2(\alpha)
\end{equation}

\bigskip

\begin{equation}\label{equ_app_datap_regularization_loss_derivative}
\frac{\partial \mathcal{L}_{\alpha_{reg}}}{\partial \alpha} = \frac{\alpha-\sigma(\alpha)}{\vert \alpha-\sigma(\alpha) \vert} (1-\sigma(\alpha)+\sigma^2(\alpha))
\end{equation}


\subsection{DAWEP Derivation}\label{app_dawep_derivation}

Derivation of DAWEP is based on cross entropy loss function explained in Section \ref{subsub_vqa_evalutaion} with Eq. \ref{equ_vqa_individual}, \ref{equ_vqa_weighted_each}, \ref{equ_vqa_weighted} and \ref{equ_vqa_final}. DAWEP is basically the gradient of each VQA answer class w.r.t to cross entropy loss as shown in Eq. \ref{equ_DAWEP_gradients}. Using the derivative product rule given in Eq. \ref{equ_app_derivative_product_rule}  where $f(x)$ and $g(x)$ are two differentiable functions, derivate of Eq. \ref{equ_vqa_final} w.r.t class weights is given in Eq. \ref{equ_app_DAWEP_derivative_partial}:

\begin{equation}\label{equ_app_derivative_product_rule}
\frac{\partial (f(x)g(x))}{\partial x} = f'(x)g(x) + f(x)g'(x)
\end{equation}

\bigskip

\begin{align}\label{equ_app_DAWEP_derivative_partial}
\mathcal{L}_{VQA}(x,y) =  \frac{\partial}{\partial w_n} \left(  \frac{1}{\sum_{n=1}^N \omega_{y_n}} \right) \left( \sum_{n=1}^N-\omega_{y_n}\log\frac{e^{x_{n,y_n}}}{\sum_{c=1}^C e^{x_{n,c}}} \right) + \nonumber \\
\left( \frac{1}{\sum_{n=1}^N \omega_{y_n}} \right)  \frac{\partial}{\partial w_n} \left( \sum_{n=1}^N -\omega_{y_n}\log\frac{e^{x_{n,y_n}}}{\sum_{c=1}^C e^{x_{n,c}}} \right)
\end{align}

\bigskip

Rest of the derivation is best explained with an example. Assume that we have batch of samples such that class weights $w_1$, $w_1$, $w_2$ along with the probabilities (obtained after softmax) $P_{1,1}$, $P_{2,1}$ and $P_{1,2}$  where $P_{i,j}$ denotes $i^{th}$ sample of $j^{th}$ class in that batch. We can directly use probabilities for simplicity since softmax function doesn't contain any terms relating to class weights. Putting these samples and weights into cross entropy loss, Eq. \ref{equ_app_samples_loss} is obtained:  

\begin{equation}\label{equ_app_samples_loss}
\mathcal{L}_{VQA}(x,y) =  \frac{1}{2w_1+w_2}(-w_1 log(P_{1,1}) -w_1 log(P_{2,1}) -w_2 log(P_{2,1}))
\end{equation}    

Taking the derivative of Eq. \ref{equ_app_samples_loss} w.r.t $w_1$ with product rule given in Eq. \ref{equ_app_derivative_product_rule}, we get Eq. \ref{equ_app_samples_loss_derivative}: 

\begin{align}\label{equ_app_samples_loss_derivative}
\frac{\partial \mathcal{L}_{VQA}}{\partial w_1} = \frac{\partial}{\partial w_1} \left(  \frac{1}{2w_1+w_2} \right) (-w_1 log(P_{1,1}) -w_1 log(P_{2,1}) -w_2 log(P_{2,1})) + \nonumber \\
 \left( \frac{1}{2w_1+w_2} \right) \frac{\partial}{\partial w_1} (-w_1 log(P_{1,1}) -w_1 log(P_{2,1}) -w_2 log(P_{2,1}))
\end{align}

Where: 

\begin{align}
\frac{\partial}{\partial w_1} \left( \frac{1}{2w_1+w_2} \right) = \frac{\partial}{\partial w_1}(2w_1+w_2)^{-1} = (-1)(2)(2w_1+w_1)^{-2} \nonumber \\
\frac{\partial}{\partial w_1} \left( \frac{1}{2w_1+w_2} \right) = - \frac{2}{(2w_1+w_2)^{2}}
\end{align}

And:

\begin{align} 
\frac{\partial}{\partial w_1}(-w_1 log(P_{1,1}) -w_1 log(P_{2,1}) -w_2 log(P_{2,1})) = -log(P_{1,1}) log(P_{2,1})
\end{align}

Putting everything together we get: 

\begin{align}\label{equ_app_samples_final}
\frac{\partial \mathcal{L}_{VQA}}{\partial w_1} =  \left( - \frac{2}{(2w_1+w_2)^{2}} \right) (-w_1 log(P_{1,1}) -w_1 log(P_{2,1}) -w_2 log(P_{2,1})) + \nonumber \\ \left( \frac{1}{2w_1+w_2} \right) (-log(P_{1,1}) - log(P_{2,1})) 
\end{align}

As can be seen from Eq. \ref{equ_app_samples_final}, first parenthesis term of the is the number of occurrences for class samples $w_1$ divided by the square of the sum of all weights. The second parenthesis term is the weighted negative log-likelihood of sample probabilities. The third parenthesis term is the normalizing term used in the cross-entropy loss function. Finally, the fourth parenthesis term is the negative log-likelihood of sample probabilities without weights. Generalizing all the derivations above, we can write Eq. \ref{equ_app_DAWEP_derivative} (also given in Eq. \ref{equ_DAWEP_derivative}) as given below:

\begin{align}\label{equ_app_DAWEP_derivative}
\frac{\partial \mathcal{L}_{VQA}}{\partial w_{\hat{n}}} = \left( -\frac{\hat{n}_{occurance}}{(\sum_{n=1}^N \omega_{y_n})^2}
\sum_{n=1}^N \omega_{y_n}\log\frac{e^{x_{n,y_n}}}{\sum_{c=1}^C e^{x_{n,c}}} \right) + \nonumber \\
\left( \frac{1}{\sum_{n=1}^N \omega_{y_n}}(\log\frac{e^{x_{y_{\hat{n}}}}}{\sum_{c=1}^{\hat{C}} e^{x_{n,c}}}) \right)
\end{align}

\section{Example Predictions (Multimodal)}\label{sec_app_example_predictions_multimodal}

\begin{figure}[H]%
\centering
\includegraphics[width=\textwidth,height=\textheight,keepaspectratio]{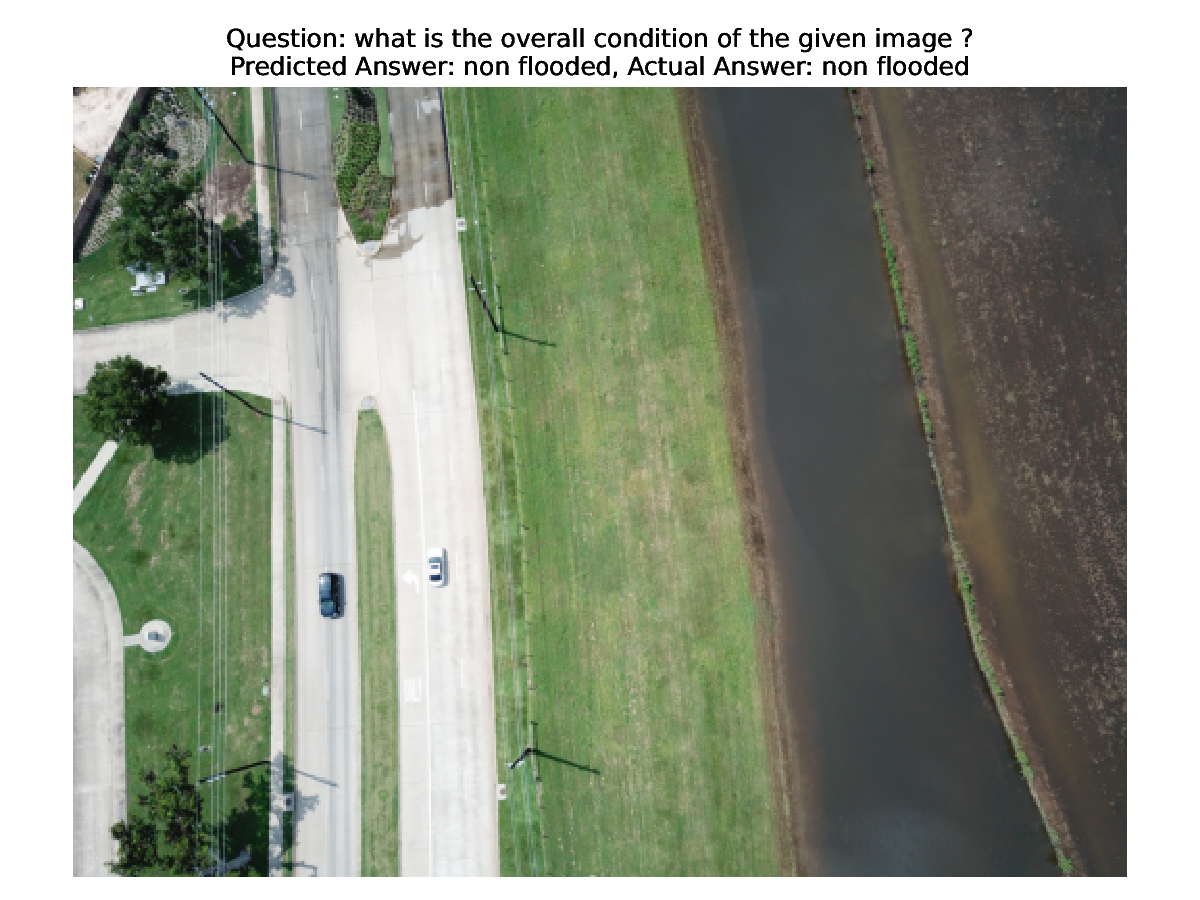}
\caption{VQA answer predictions for "Condition Recognition" with image and question sentence}\label{fig_vqa_pair_1}
\end{figure}

\begin{figure}[H]%
\centering
\includegraphics[width=0.95\textwidth]{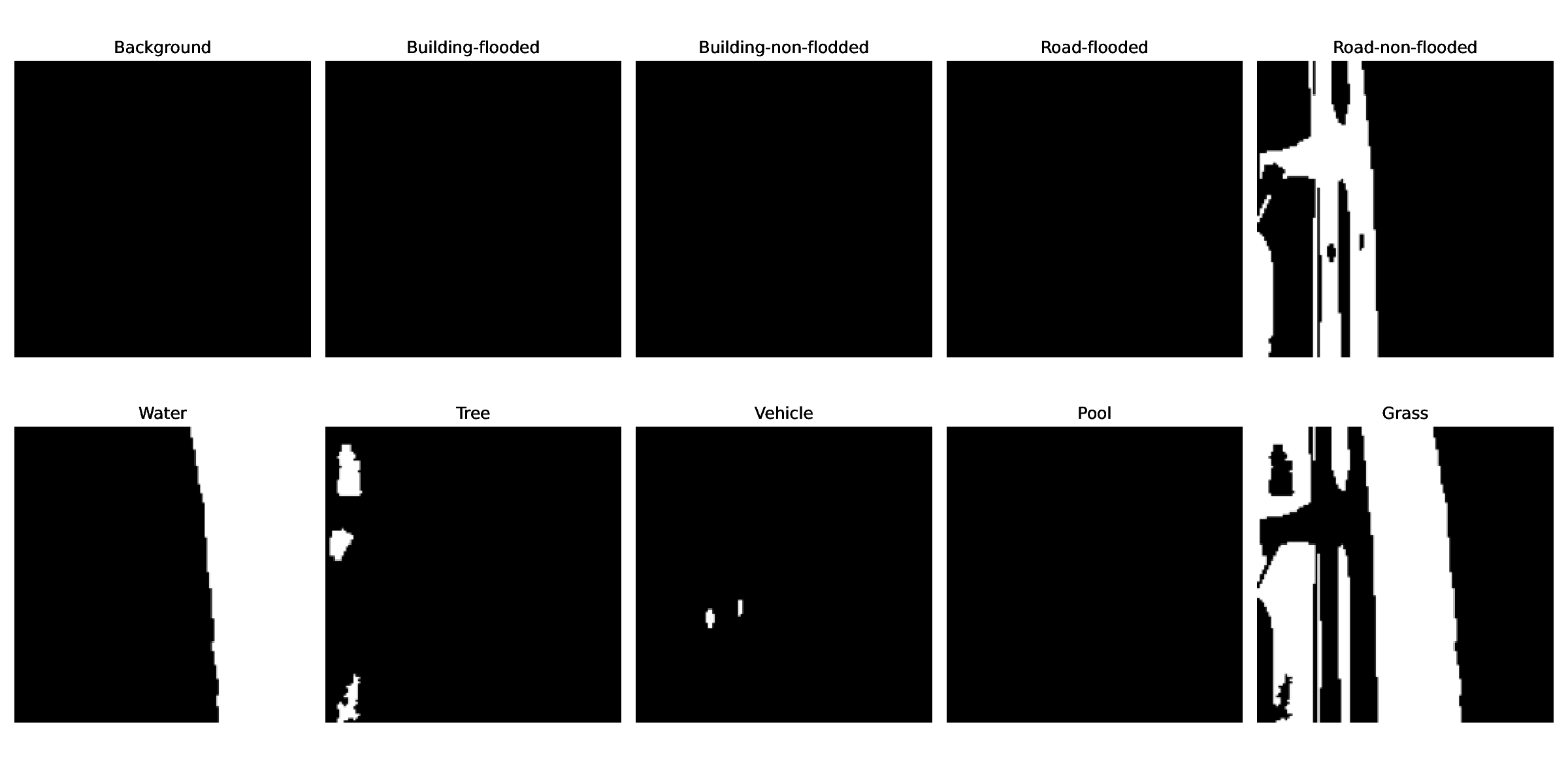}
\caption{Segmentation mask ground truths from FloodNet dataset for each class.}\label{fig_masks_actual}
\end{figure}

\begin{figure}[H]%
\centering
\includegraphics[width=0.95\textwidth]{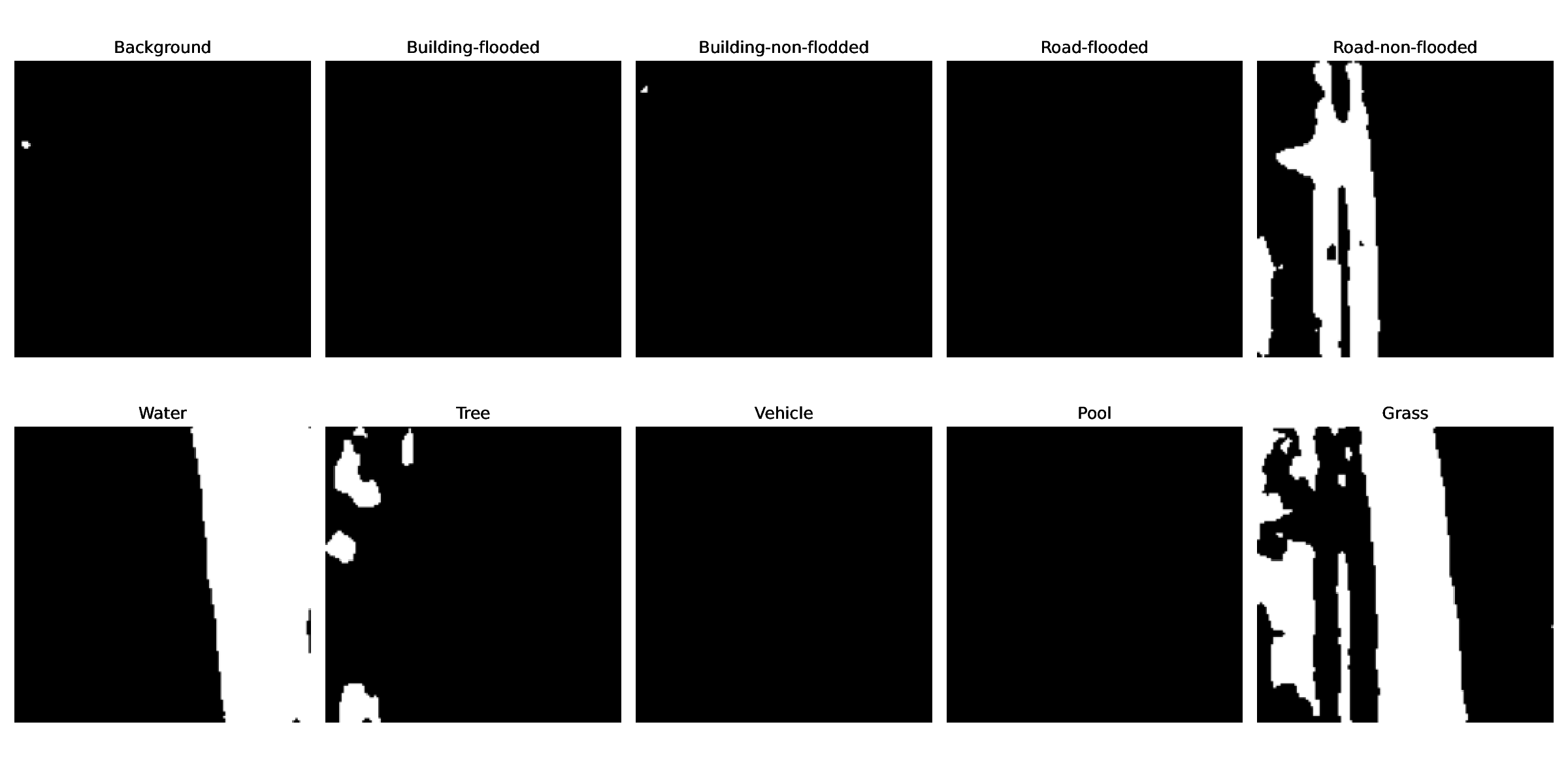}
\caption{Segmentation mask predictions of our deep neural network for each class.}\label{fig_masks_prediction}
\end{figure}

\section{Example Predictions (VQA Only)}\label{sec_app_example_predictions_vqa_only}

\begin{figure}[H]%
\centering
\includegraphics[width=\textwidth,height=\textheight,keepaspectratio]{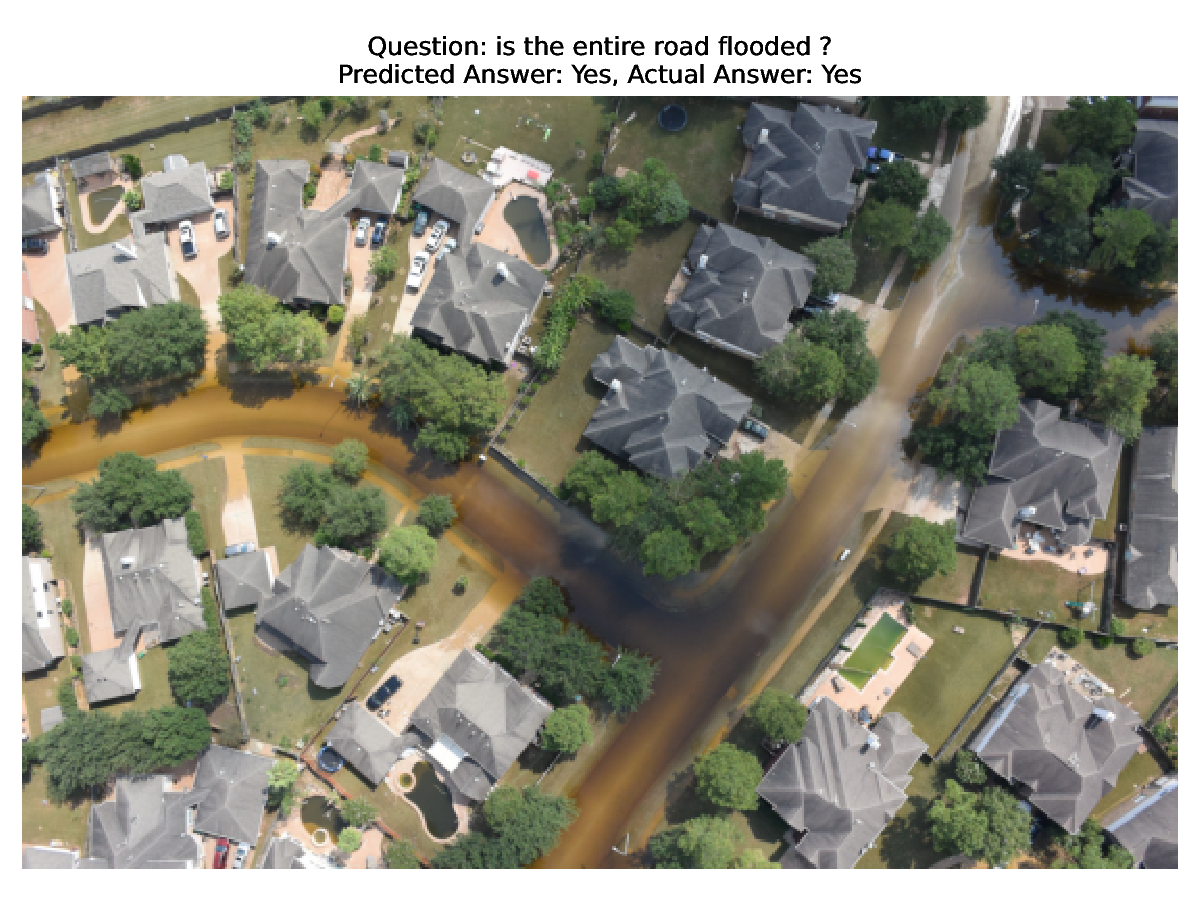}
\caption{VVQA answer predictions for "Yes/No" question type}\label{fig_vqa_pair_2}
\end{figure}

\begin{figure}[H]%
\centering
\includegraphics[width=\textwidth,height=\textheight,keepaspectratio]{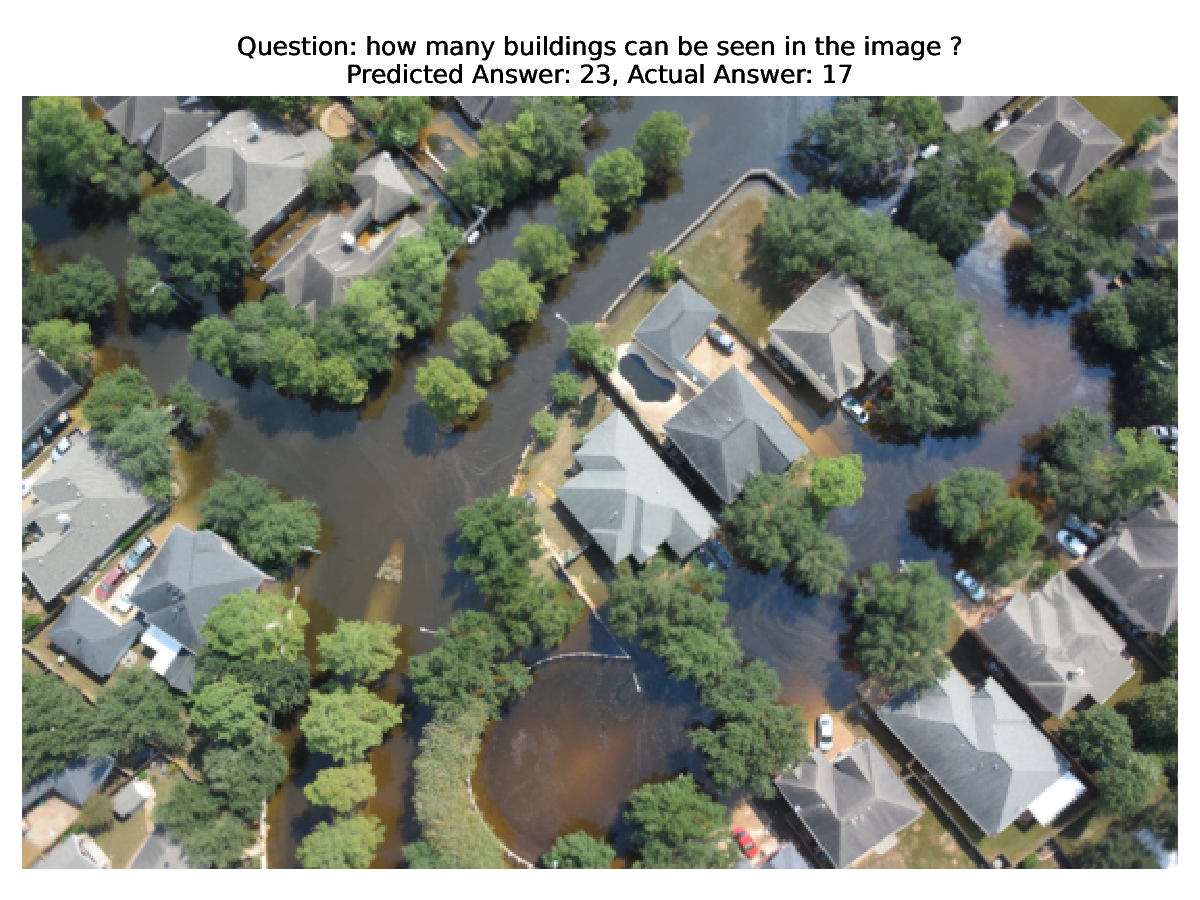}
\caption{VQA answer predictions for "Simple Counting" question type}\label{fig_vqa_pair_3}
\end{figure}

\begin{figure}[H]%
\centering
\includegraphics[width=\textwidth,height=\textheight,keepaspectratio]{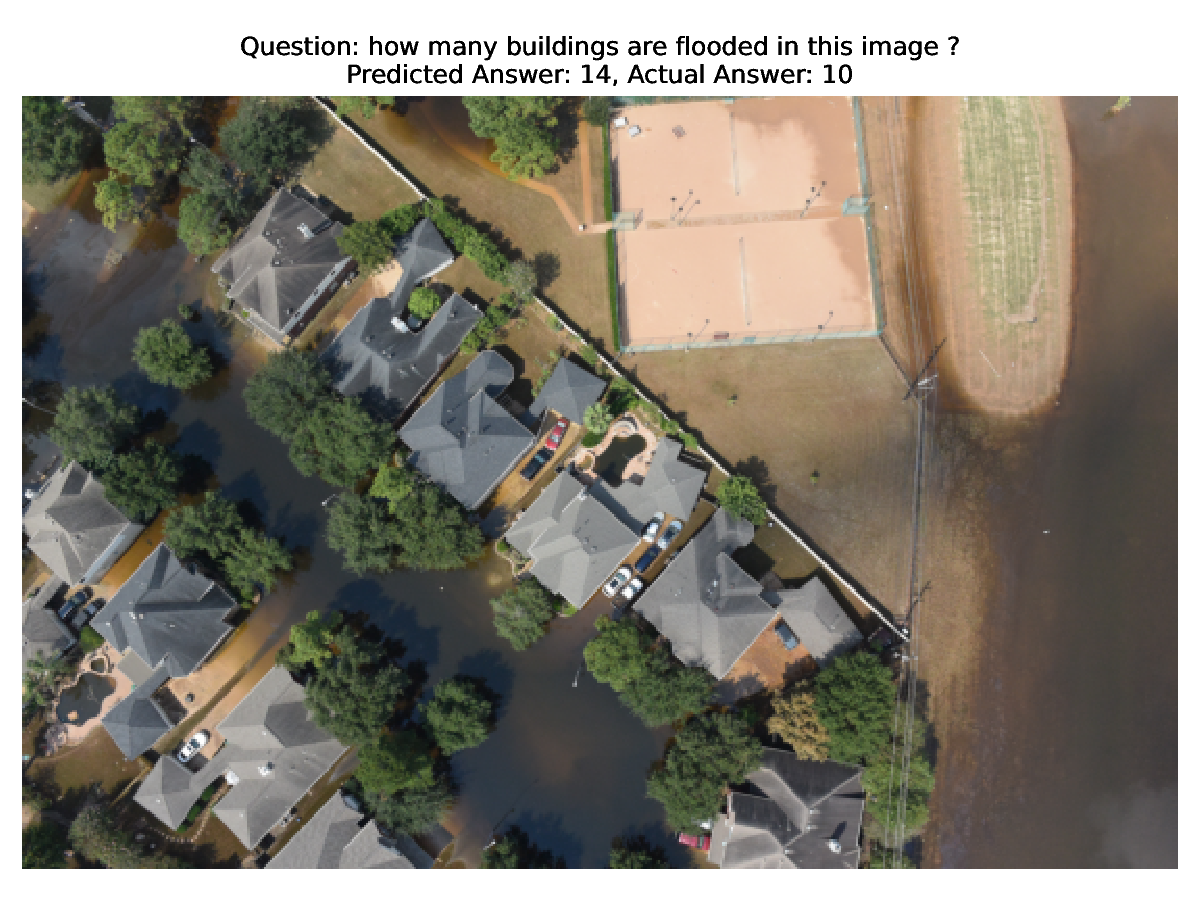}
\caption{VQA answer predictions for "Complex Counting" question type}\label{fig_vqa_pair_4}
\end{figure}

\bibliography{references}

\end{document}